\documentclass[10pt,letterpaper,twoside,onecolumn,final]{article}
\usepackage{ecj,palatino,latexsym,natbib}
\usepackage[ruled,vlined,linesnumbered]{algorithm2e}
\usepackage{subfigure,multirow,gensymb}
\usepackage{graphicx}
\usepackage{gnuplot-lua-tikz}
\usepackage{pgfplots}
\newcommand{\figwidth}{0.90in}
\newcommand{\rotationwidth}{2.25cm}
\newcommand{\figscale}{0.5}
\newcommand{\graphwidth}{0.81\textwidth}
\newcommand{\graphheight}{0.5\textwidth}
\newcommand{\figfontsize}{\small}
\usetikzlibrary{arrows,decorations.pathmorphing,decorations.markings,backgrounds,positioning,fit,petri}
\graphicspath{{figs/}}
\begin{document}

\ecjHeader{x}{x}{xxx-xxx}{2015}{Design Mining Interacting Wind Turbines}{R. J. Preen and L. Bull}
\title{\bf Design Mining Interacting Wind Turbines}  

\author{\name{\bf Richard J. Preen} \hfill \addr{richard2.preen@uwe.ac.uk}\\ 
\addr{Department of Computer Science and Creative Technologies,\\ University of the West of England, Bristol, BS16 1QY,\\ United Kingdom}
\AND
\name{\bf Larry Bull} \hfill \addr{larry.bull@uwe.ac.uk}\\
\addr{Department of Computer Science and Creative Technologies,\\ University of the West of England, Bristol, BS16 1QY,\\ United Kingdom} }

\maketitle

\begin{abstract}

An initial study of surrogate-assisted evolutionary algorithms used to design vertical-axis wind turbines wherein candidate prototypes are evaluated under fan generated wind conditions after being physically instantiated by a 3D printer has recently been presented. Unlike other approaches, such as computational fluid dynamics simulations, no mathematical formulations were used and no model assumptions were made. This paper extends that work by exploring alternative surrogate modelling and evolutionary techniques. The accuracy of various modelling algorithms used to estimate the fitness of evaluated individuals from the initial experiments is compared. The effect of temporally windowing surrogate model training samples is explored. A surrogate-assisted approach based on an enhanced local search is introduced; and alternative coevolution collaboration schemes are examined.
\end{abstract}


\section{Introduction}
  
Renewable energy contributed over half of total net additions to global electric generating capacity from all sources in 2012, with wind power accounting for around 39\% of the renewable power added~\citep[p.~13]{Renewables:2013}. Currently, arrays of horizontal-axis wind turbines (HAWTs) are the most commonly used form of wind farm employed to extract large amounts of wind energy. However, as the turbines extract the energy from the wind, the energy content decreases and the amount of turbulence increases downstream from each. For example, see \cite{Hasager:2013} for photographs and explanation of the well-known wake effect at the Horns Rev offshore wind farm in the North Sea. Due to this, HAWTs must be spaced 3--5 turbine diameters apart in the cross-wind direction and 6--10 diameters apart in the downwind direction in order to maintain 90\% of the performance of isolated HAWTs~\citep{Dabiri:2011}. The study of these wake effects is therefore a very complex and important area of research~\citep{Barthelmie:2006}, as is turbine placement~\citep{Mosetti:1994}. Thus, ``modern wind farms comprised of HAWTs require significant land resources to separate each wind turbine from the adjacent turbine wakes. This aerodynamic constraint limits the amount of power that can be extracted from a given wind farm footprint. The resulting inefficiency of HAWT farms is currently compensated by using taller wind turbines to access greater wind resources at high altitudes, but this solution comes at the expense of higher engineering costs and greater visual, acoustic, radar and environmental impact''~\citep[p.~1]{Dabiri:2011}. This has forced wind energy systems away from high energy demand population centres and toward remote locations with higher distribution costs. 

In contrast, vertical-axis wind turbines (VAWTs) do not need to be oriented to wind direction and the spacing constraints of HAWTs often do not apply. VAWT performance can even be increased by the exploitation of inter-turbine flow effects~\citep{Charwat:1978}. In addition, VAWTs can also be easier to manufacture, may scale more easily, are typically inherently light-weight with little or no noise pollution, and are more able to tolerate extreme weather conditions~\citep{Eriksson:2008}. This has resulted in a recent expansion of their use in urban environments~\citep{Toja-Silva:2013}. However, their design space is complex and relatively unexplored. Generally, two classes of design are currently under investigation and exploitation: \cite{Savonius:1930}, which has blades attached directly upon the central axis structure; and \cite{Darrieus:1931}, where the blades---either straight or curved---are positioned predominantly away from the central structure. Hybrids also exist. The small body of previous work considering VAWT farms/arrays has used turbines originally intended to operate alone. Our work is the only known work to consider designing VAWT in arrays of interacting turbines.

We have recently presented an initial study~\citep{PreenBull:2014a} of surrogate-assisted genetic algorithms \citep[SGAs;][]{Dunham:1963} used to design VAWTs wherein candidate prototypes are evaluated under fan generated wind conditions after being physically instantiated by a 3D printer. That is, unlike other approaches, no mathematical formulations are used and no model assumptions are made. Initially, artificial evolution was used to explore the design space of a single isolated VAWT and subsequently a cooperative coevolutionary genetic algorithm \citep[CGA;][]{HusbandsMill:1991} was applied to explore the design space of an array of 2 closely positioned VAWTs. Both conventional CGA and surrogate-assisted (SCGA) versions were examined, finding increased aerodynamic performance (rotational speed) in fewer fabrications with surrogate assistance. For single turbine comparison, the fittest evolved designs were found to be aerodynamically more efficient than several common human designs under the experimental conditions. In this paper, we extend that work by exploring alternative surrogate modelling and evolutionary techniques. First, the accuracy of various modelling techniques used to estimate the fitness of the individuals from the initial experiments is compared. Subsequently, we compare surrogate model performance with different training samples. An alternative surrogate approach based on an enhanced local search is introduced. Finally, alternative coevolution collaboration schemes are examined, including one that considers the potential for symmetry within the task.

\section{Background}

\subsection{Interacting Vertical-Axis Wind Turbines}
  
\begin{figure}[t]
	\centering 
	\subfigure[Clockwise co-rotation.]{
		\begin{tikzpicture}
			\draw[red, line width=2pt, ->, >=stealth] (0:-0.8) arc (120:50:{\rotationwidth-0.5cm});
			\node [inner sep=5pt,below] {\includegraphics[width=\rotationwidth]{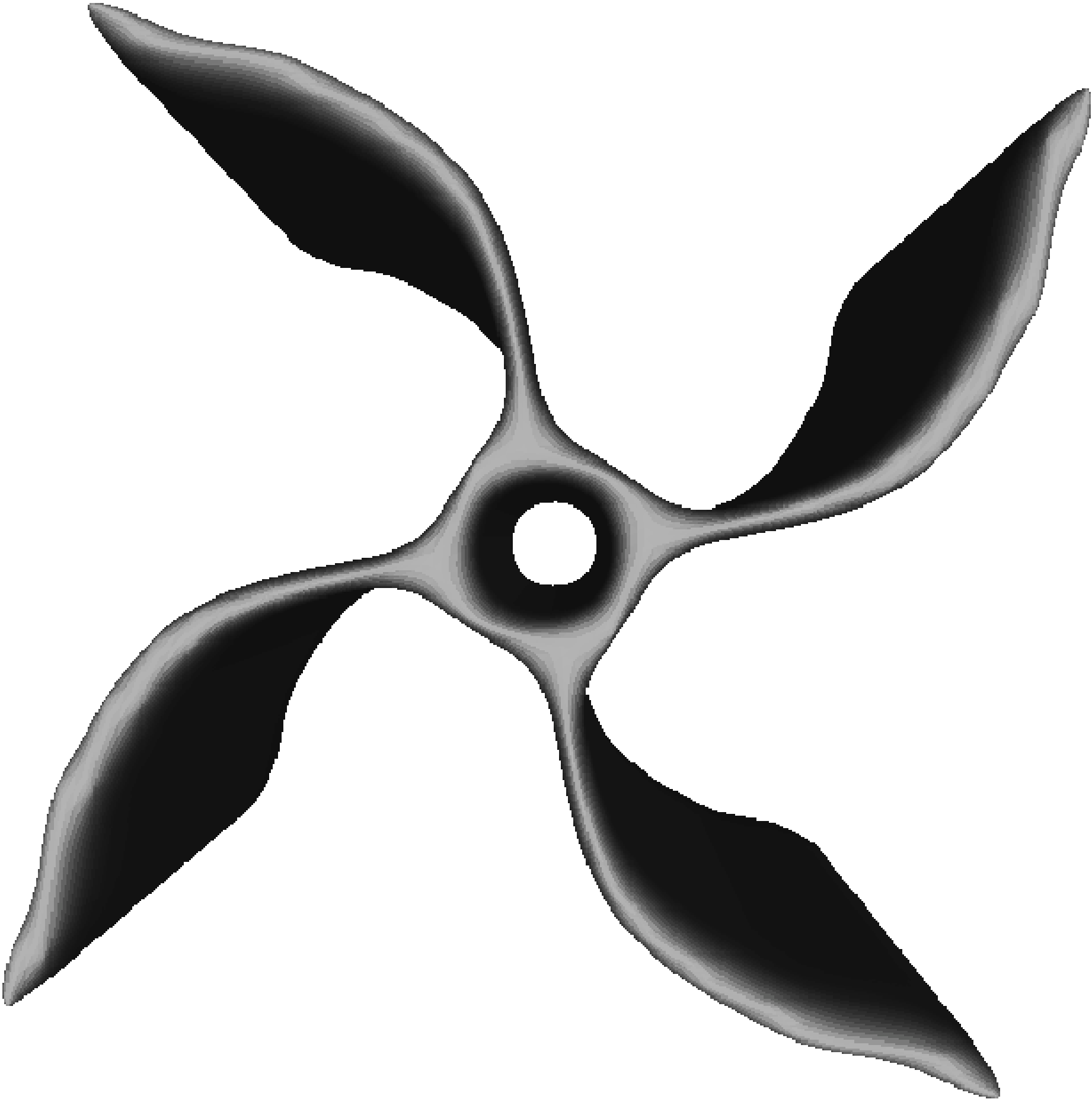}};
		\end{tikzpicture}
		\begin{tikzpicture}
			\draw[red, line width=2pt, ->, >=stealth] (0:-0.8) arc (120:50:{\rotationwidth-0.5cm});
			\node [inner sep=5pt,below] {\includegraphics[width=\rotationwidth]{fig-1.png}};
		\end{tikzpicture}
	} \hspace{0.5in}
	\subfigure[Counter-clockwise co-rotation.]{
			\scalebox{-1}[1]{
			\begin{tikzpicture}
				\draw[red, line width=2pt, ->, >=stealth] (0:-0.8) arc (120:50:{\rotationwidth-0.5cm});
				\node [inner sep=5pt,below] {\includegraphics[width=\rotationwidth]{fig-1.png}};
			\end{tikzpicture}
			\begin{tikzpicture}
				\draw[red, line width=2pt, ->, >=stealth] (0:-0.8) arc (120:50:{\rotationwidth-0.5cm});
				\node [inner sep=5pt,below] {\includegraphics[width=\rotationwidth]{fig-1.png}};
			\end{tikzpicture}
		}
	} \\
	\subfigure[Counter-rotation; with a northerly wind, the upwind blades are in the middle region.]
	{ 
	   \scalebox{-1}[1]{
	   \begin{tikzpicture}
			\draw[red, line width=2pt, ->, >=stealth] (0:-0.8) arc (120:50:{\rotationwidth-0.5cm});
			\node [inner sep=5pt,below] {\includegraphics[width=\rotationwidth]{fig-1.png}};
		\end{tikzpicture}}
		\begin{tikzpicture}
			\draw[red, line width=2pt, ->, >=stealth] (0:-0.8) arc (120:50:{\rotationwidth-0.5cm});
			\node [inner sep=5pt,below] {\includegraphics[width=\rotationwidth]{fig-1.png}};
		\end{tikzpicture}
	} \hspace{0.5in}
	\subfigure[Counter-rotation; with a northerly wind, the driving blades are in the middle region.]
	{ 
	   \begin{tikzpicture}
			\draw[red, line width=2pt, ->, >=stealth] (0:-0.8) arc (120:50:{\rotationwidth-0.5cm});
			\node [inner sep=5pt,below] {\includegraphics[width=\rotationwidth]{fig-1.png}};
		\end{tikzpicture}
		\scalebox{-1}[1]{
		\begin{tikzpicture}
			\draw[red, line width=2pt, ->, >=stealth] (0:-0.8) arc (120:50:{\rotationwidth-0.5cm});
			\node [inner sep=5pt,below] {\includegraphics[width=\rotationwidth]{fig-1.png}};
	\end{tikzpicture}}
	}
	\caption{Possible VAWT pair rotation configurations.}
	\label{fig:rotations}
\end{figure}

Arrays of closely spaced VAWTs have long been considered for use as wind power stations. For example, \cite{Charwat:1978} observed an improvement in the performance of a pair of closely spaced `S' shaped VAWT (S-rotors), whether counter or co-rotating, compared with that of a single turbine. Figure~\ref{fig:rotations} illustrates the possible rotation configurations of a pair of VAWT. Despite this, VAWTs have been restricted to niche applications since a single HAWT provides a much higher efficiency compared with a single VAWT. However, recently \cite{Kinzel:2012} performed an experimental field study of an array of 9 pairs of full-scale counter-rotating VAWTs. They found that the wind velocity behind a turbine pair recovers to 95\% of the wind velocity upwind after approximately 6 turbine diameters, compared with 4 diameters for the wake behind a single VAWT and 14 diameters for HAWTs. Thus, closely spaced VAWTs can result in an overall reduction in the average inter-turbine spacing as well as increasing individual performance, leading to a much greater power density. Indeed, it has recently been shown~\citep{Dabiri:2011} that power densities an order of magnitude greater can be potentially achieved by arranging VAWTs in layouts utilising counter-rotation that enable them to extract energy from adjacent wakes and from above the wind farm.                

A growing body of work has been exploring techniques to optimise a given wind farm layout---termed micro-siting. Algorithms used include, genetic algorithms~\citep[GAs;][]{Mosetti:1994}, evolution strategies~\citep[ESs;][]{KusiakSong:2010}, particle swarm optimisation~\citep{Aristidis:2010}, ant colony optimisation~\citep{Eroglu:2012}, Monte Carlo simulation~\citep{Marmidis:2008}, and principles from fish schooling~\citep{Whittlesey:2010}. See \cite{Salcedo-Sanz:2011} for a recent review of evolutionary computation-based techniques for micro-siting. Importantly, all of this work has been based on wake models of varying degrees of fidelity.

To date, heterogeneity in wind farms has been almost completely unexplored. \cite{Chamorro:2014} recently investigated a variable size HAWT array composed of $3\times8$ model wind turbines where large and small turbines were alternately positioned. They found that size heterogeneity has positive effects on turbulent loading as a result of the larger turbines facing a more uniform turbulence distribution and the smaller turbines operating under lower turbulence levels. The interactions show the possibility that heterogeneity within wind farms has the potential to improve the overall ability to harvest energy. Our initial study~\citep{PreenBull:2014a} observed that VAWT array asymmetry can be more efficient than similar symmetrical designs; for example, the individuals from the fittest SCGA array pairing after 160 fabrications were duplicated to form homogeneous arrays and the maximum combined rotational speed was found to be lower than the heterogeneous array.
  
\subsection{Evolutionary Design}

Evolutionary algorithms (EAs) have long been used to design 3D physical objects. Notably, \cite{Hornby:2011} evolved and manufactured an {X}-band satellite antenna for NASA's ST5 spacecraft, representing the world's first artificially evolved hardware in space. Significantly, the evolved antennas outperformed a design hand-produced by the antenna contractor for the mission. Most approaches, however, have used simulations to provide the fitness scores of the evolved designs before final fabrication.
  
The majority of blade design optimisation is performed through the use of CFD simulations, typically described with 3D Navier-Stokes equations~\citep{Anderson:1995}. However, 3D CFD simulations are computationally expensive, with a single calculation taking hours on a high-performance computer, making their use with an iterative search approach difficult. Moreover, assumptions need to be made, e.g.,\ regarding turbulence or pressure distributions, which can significantly affect accuracy when modelling interacting wind turbines. Previous evolutionary studies have been undertaken with types of CFD to optimise the blade profile for both HAWT~\citep{Hampsey:2002} and VAWT~\citep{Carrigan:2012} to varying degrees of success/realism.

The evaluation of physical artifacts directly for fitness determination can be traced back to the origins of evolutionary computation~\citep{Dunham:1963}. For example, the first ESs were used to design jet nozzles with a string of section diameters, which were then machined and tested for fitness~\citep{Rechenberg:1971}. Other well-known examples include robot controller design~\citep{Nolfi:1992}, electronic circuit design using programmable hardware~\citep{Thompson:1998}, product design via human provided fitness values~\citep{Herdy:1996}, chemical systems~\citep{Theis:2006}, and unconventional computers~\citep{HardingMiller:2004}. More recently, \cite{Boria:2009} used an EA to evolve a morphing wing structure where physical designs were morphed using a set of actuators and evaluated in a closed-loop wind tunnel. 

Evolution in hardware has the potential to benefit from access to a richer environment where it can exploit subtle interactions that can be utilised in unexpected ways. For example, the EA used by \cite{Thompson:1998} to work with field-programmable gate array circuits used some subtle physical properties of the system to solve problems where the properties used are still not understood. Humans can be prevented from designing systems that exploit these subtle and complex physical characteristics through their lack of knowledge, however this does not prevent exploitation through artificial evolution. There is thus a real possibility that evolution in hardware may allow the discovery of new physical effects, which can be harnessed for computation/optimisation~\citep{MillerDowning:2002}.  

Moreover, the advent of high quality, low-cost, additive rapid fabrication technology---known as 3D printing---means it is now possible to fabricate a wide range of prototype designs quickly and cheaply. 3D printers are now capable of printing an ever growing array of different materials, including food, e.g.,\ chocolate~\citep{Hao:2009} and meat~\citep{Lipton:2010} for culinary design; sugar, e.g.,\ to help create synthetic livers~\citep{Miller:2012}; chemicals, e.g.,\ for custom drug design~\citep{Cronin:2012}; cells, e.g.,\ for functional blood vessels~\citep{Jakeb:2008} and artificial cartilage~\citep{Xu:2013}; plastic, e.g.,\ Southampton University laser sintered aircraft; thermoplastic, e.g.,\ for electronic sensors~\citep{Leigh:2012}; titanium, e.g.,\ for prosthetics such as the synthetic mandible developed by the University of Hasselt and transplanted into an 83-year old woman; and liquid metal, e.g.,\ for stretchable electronics~\citep{Ladd:2013}. One potential benefit of the technology is the ability to perform fabrication directly in the target environment; for example, \cite{Cohen:2010} recently used a 3D printer to perform a minimally invasive repair of the cartilage and bone of a calf femur {\it in situ}. \cite{LipsonPollack:2000} were the first to exploit the emerging technology in conjunction with an EA using a simulation of the mechanics and control, ultimately printing mobile robots with embodied neural network controllers.

\subsection{Surrogate-Assisted Evolutionary Algorithms}
 
Whilst the speed and cost of rapid-prototyping continues to improve, fabricating an evolved design before fitness can be assigned remains an expensive task when potentially thousands of evaluations are required; e.g.,\ 10~minutes print time for each very simple individual in the work of~\cite{RieffelSayles:2010}. However, given a sample $\mathcal{D}$ of evaluated individuals $N$, a surrogate model (also known as a meta-model or response surface model) $y=f(\vec{x})$ can be constructed, where $\vec{x}$ is the genotype and $y$ fitness, in order to compute the fitness of an unseen data point $\vec{x} \notin \mathcal{D}$. The use of surrogate models has been shown to reduce the convergence time in evolutionary computation and multiobjective optimisation; see~\cite{Jin:2011,Viana:2014} for recent general reviews and \cite{ForresterKeane:2009} for aerospace design optimisation.

Typically, a set of evaluated genotypes and their real fitness scores are used to perform the supervised training of a multi-layer perceptron \citep[MLP;][]{Rosenblatt:1962} based artificial neural network \citep[e.g.,][]{Eberhart:1992}. However, other approaches are widely used, for example, kriging~\citep[e.g.,][]{Ratle:2001}, clustering~\citep[e.g.,][]{KimCho:2001}, support vector regression~\citep[e.g.,][]{Yun:2009}, radial-basis functions~\citep[e.g.,][]{Ong:2006}, and sequential parameter optimisation~\citep[e.g.,][]{Bartz-Beielstein:2006}. The surrogate model is subsequently used to compute estimated fitness values for the EA to utilise. The model must be periodically retrained with new individuals under a controlled evolutionary approach (also known as model management) to prevent convergence on local optima. Retraining can be performed by taking either an individual or generational approach. In the individual approach, $n$ number of individuals in the population, $P$, are chosen and evaluated with the real fitness function each generation, and in the generational approach the entirety of $P$ is evaluated on the real fitness function each $n$-th generation. Typically the members with the highest approximated fitness are chosen to be evaluated with the real fitness function in the individual approach, although alternative schemes have been suggested; \cite{Bull:1999} found that evaluating both the best and random individuals suggested by a neural network surrogate model resulted in a significant improvement over exclusively evaluating either the best or random individuals. Both global modelling and local modelling using trust regions (e.g.,\ samples within a certain Euclidean distance) are popular approaches~\citep{Le:2013}. Resampling methods and surrogate model validation remain an important and ongoing area of research, enabling the comparison and optimisation of models~\citep{Bischl:2012}.

The use of approximations in a coevolutionary context has previously been shown capable of solving computationally expensive optimisation problems with varying degrees of epistasis more efficiently than conventional CGAs through the use of radial basis functions~\citep{Ong:2002} and memetic algorithms~\citep{Goh:2011}.

\subsection{Coevolution}
 
Since the early work in evolving both competitive~\citep{Axelrod:1987} and cooperative~\citep{HusbandsMill:1991} multi-agent systems, the problem of how to pick evaluation partners has been noted. Many strategies have been presented, which vary from the extreme case of each individual in one population using all others in all other populations~\citep[e.g.,][]{Koza:1991}, to each individual using a subset of the others~\citep[e.g.,][]{Hillis:1991}, to the more computationally preferable use of one individual from each population~\citep[e.g.,][]{BullFogarty:1993}. The use of the current best individuals from the other species populations was examined by \cite{PotterDeJong:1994}. In their work, all populations also received a shared fitness measure. Using a generational CGA they reported that the strategy performed well. They also suggested that for higher cross-species epistasis an additional randomly selected individual should be used as a partner and the highest obtained fitness from the two pairings assigned to the evaluating genome. \cite{Bull:1997b} examined various strategies for choosing evaluation partners from coevolving populations and also found improved performance on problems with significant cross-species epistasis when using multi-partner strategies. No significant difference was observed whether the second collaborator was chosen randomly or fitness proportionately after~\cite{Paredis:1994}. These early results suggested that the collaboration method depends entirely upon the amount of epistatic interaction between different species, however the issue is more complex; for example, \cite{Wiegand:2001} found that simply using the fittest individual as the collaborating partner worked best on a non-linearly separable quadratic problem with cross-species epistasis. In addition, they found that when using multiple collaborative function evaluations, assigning the fitness score from the best collaboration is significantly better than using either the worst or average of the scores.

\section{Methodology}

Here, we investigate the use of SCGAs to design small wind farms, utilising the aggregated rotational speed of the array as fitness. Each VAWT is treated separately by evolution and approximation techniques, i.e.,\ heterogeneous designs could emerge.

A vector of 10 integers is used as a simple and compact encoding of the $x$-$y$-axis of a prototype VAWT. Each allele thus controls 1/10th of a single $z$-layer. A workspace (maximum object size) of $30\times30\times30$~mm is used so that the instantiated prototype is small enough for timely production ($\sim30$~minutes) and with low material cost, yet large enough to be sufficient for fitness evaluation. The workspace has a resolution of $100\times100\times100$ voxels. A central platform is constructed for each individual to enable the object to be placed on to the evaluation equipment. The platform consists of a square torus, 1 voxel in width and with a centre of $14\times14$ empty voxels that are duplicated for each $z$ layer, thus creating a hollow tube that is 3~mm in diameter.

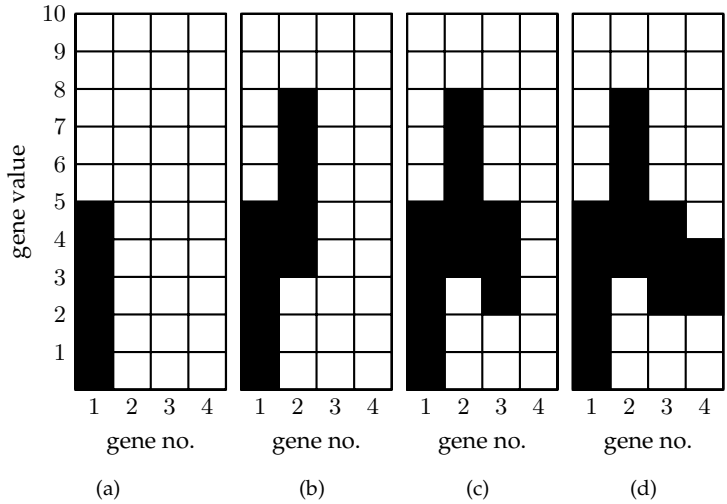
\begin{figure}[t]
	\centering
		\subfigure[]{\label{fig:d1}
			\begin{tikzpicture}[scale=\figscale, font=\figfontsize] 
				\draw [thick, draw=black, fill=white] (0,0) grid (4,10) rectangle (0,0);
				\draw [thick, draw=black, fill=black] (0,0) rectangle (1,5);
				\foreach \i in {1,...,4} {
					\draw (\i,1) -- (\i,10) node [below] at (\i-0.5,0) {$\i$};
				}
				\node[] at (2,-1.5) {gene no.};
				\foreach \i in {1,...,10} {
					\draw (1,\i) -- (4,\i) node [left] at (0,\i) {$\i$};
				}
				\node[rotate=90] at (-1.5,5) {gene value};
			\end{tikzpicture}
		}%
		\subfigure[]{\label{fig:d2}
			\begin{tikzpicture}[scale=\figscale, font=\figfontsize] 
				\draw [thick, draw=black, fill=white] (0,0) grid (4,10) rectangle (0,0);
				\draw [thick, draw=black, fill=black] (0,0) rectangle (1,5);
				\draw [thick, draw=black, fill=black] (1,3) rectangle (2,8);
				\foreach \i in {1,...,4} {
					\draw (\i,1) -- (\i,10) node [below] at (\i-0.5,0) {$\i$};
				}
				\node[] at (2,-1.5) {gene no.};
			\end{tikzpicture}
		}%
		\subfigure[]{\label{fig:d3}
			\begin{tikzpicture}[scale=\figscale, font=\figfontsize] 
				\draw [thick, draw=black, fill=white] (0,0) grid (4,10) rectangle (0,0);
				\draw [thick, draw=black, fill=black] (0,0) rectangle (1,5);
				\draw [thick, draw=black, fill=black] (1,3) rectangle (2,8);
				\draw [thick, draw=black, fill=black] (2,2) rectangle (3,5);
				\foreach \i in {1,...,4} {
					\draw (\i,1) -- (\i,10) node [below] at (\i-0.5,0) {$\i$};
				}
				\node[] at (2,-1.5) {gene no.};
			\end{tikzpicture}
		}%
		\subfigure[]{\label{fig:d4}
			\begin{tikzpicture}[scale=\figscale, font=\figfontsize] 
				\draw [thick, draw=black, fill=white] (0,0) grid (4,10) rectangle (0,0);
				\draw [thick, draw=black, fill=black] (0,0) rectangle (1,5);
				\draw [thick, draw=black, fill=black] (1,3) rectangle (2,8);
				\draw [thick, draw=black, fill=black] (2,2) rectangle (3,5);
				\draw [thick, draw=black, fill=black] (3,2) rectangle (4,4);
				\foreach \i in {1,...,4} {
					\draw (\i,1) -- (\i,10) node [below] at (\i-0.5,0) {$\i$};
				}
				\node[] at (2,-1.5) {gene no.};
			\end{tikzpicture}
		}
	\caption{Translation of $x$-$y$-axis genome [5,8,2,4]. In (a) the voxels are enabled from the first allele (5) to the baseline (bottom). Subsequently in (b) the voxels are enabled from the second allele (8) to the previous allele (5) and extended 2 voxels. In (c) the third allele (2) is less than the previous lower position (3), causing a gap, and is thus drawn from the allele (2) to the previous lower position and extended 2 voxels to provide structural support. In (d) the allele (4) is less than the previous upper position (5) and 2 voxels are enabled from the allele toward the baseline. }
	\label{fig:drawing}
\end{figure}

To translate the genome for a single $z$ layer, an equilateral cross is constructed using the 10 aforementioned genes, with 4 blades bent at right angles and an allele range [1,42]. For north-east and south-west quadrants the baseline is a horizontal line at $y$-axis=50, and for north-west and south-east quadrants the baseline is a vertical line at $x$-axis=50. Starting from the central platform and translating each gene successively, the one-tenth of voxels controlled by that gene are then drawn from the allele+baseline toward the baseline; see Figure~\ref{fig:d1}. If the current allele+baseline is greater than or the same as the previous allele+baseline, the voxels are enabled from the current allele+baseline to the previous allele+baseline and extended a further 2 voxels toward the baseline for structural support; see Figure~\ref{fig:d2}. If the current allele+baseline is less than or the same as the previous lower ending position, causing a gap, the voxels are enabled from the current allele+baseline upwards to the previous lower position and extended a further 2 voxels; see Figure~\ref{fig:d3}. In all other cases, 2 voxels are enabled from the current allele+baseline position toward the baseline; see Figure~\ref{fig:d4}. 

\begin{figure}[t]
	\centering 
	\subfigure{\includegraphics[width=1.2in]{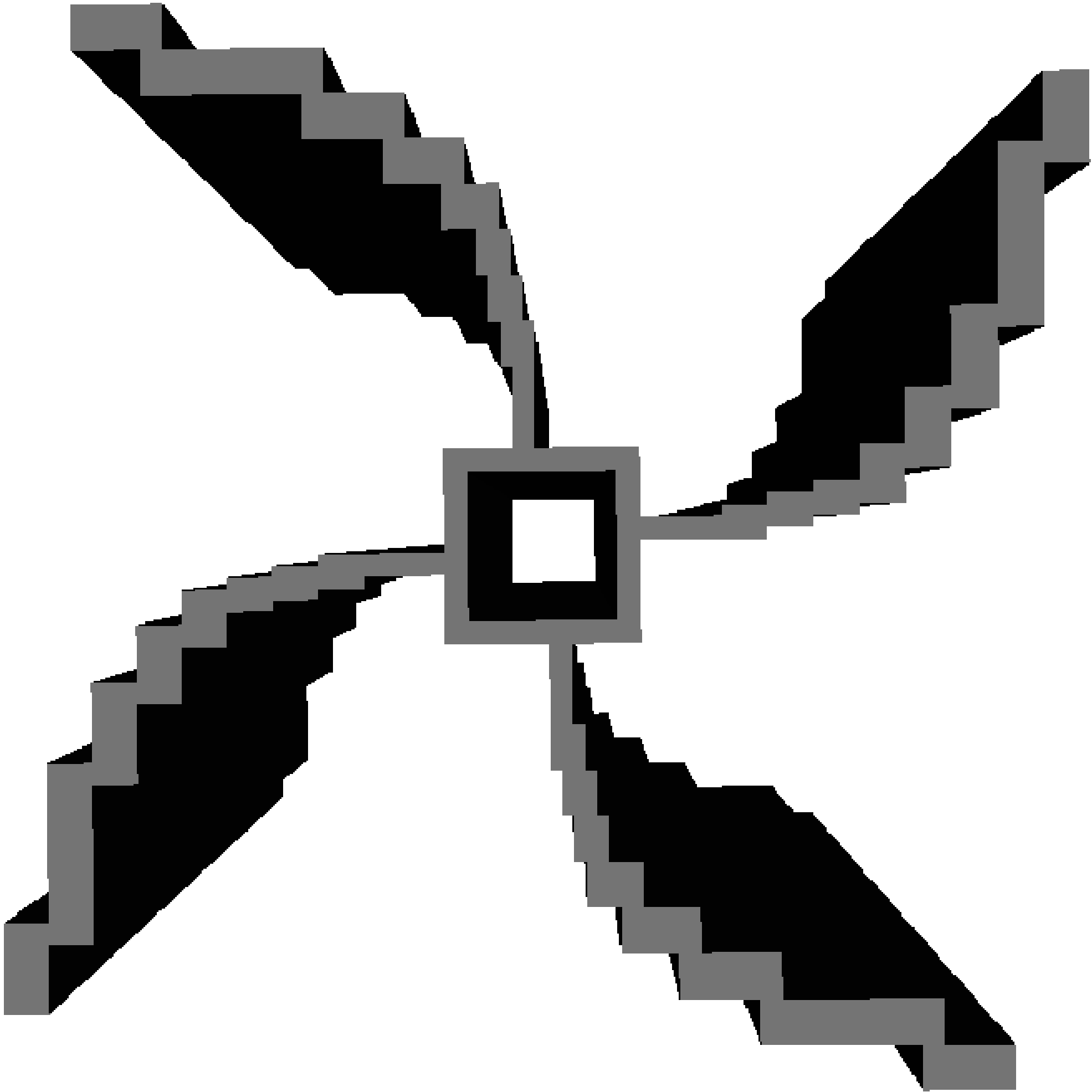} } \hspace{0.2in}
	\subfigure{\includegraphics[width=1.2in]{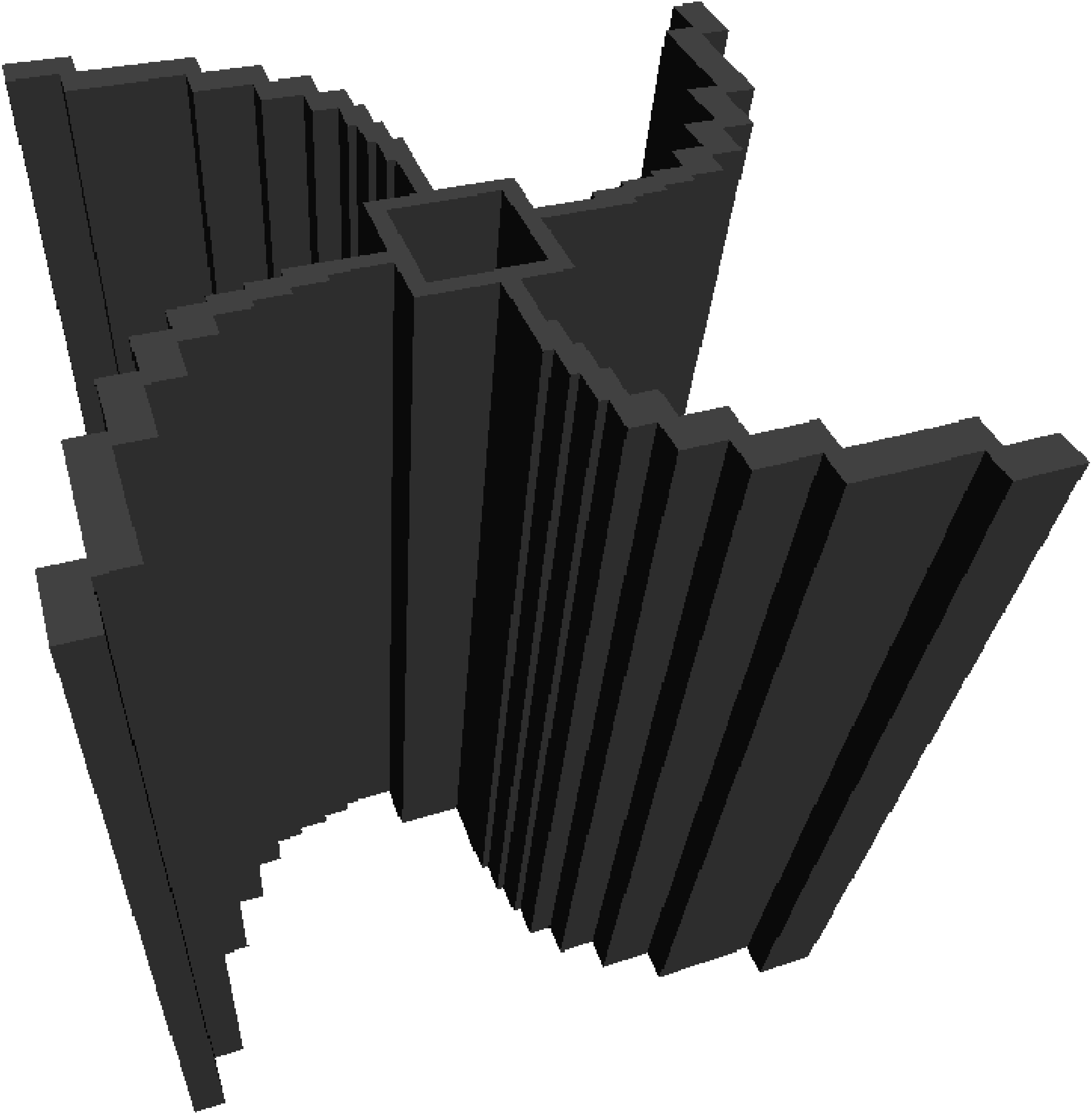} }
	\caption{Example phenotype; genome = [2, 2, 3, 4, 5, 8, 13, 20, 34, 40].}\label{fig:target-unsmoothed}
	\subfigure{\includegraphics[width=1.2in]{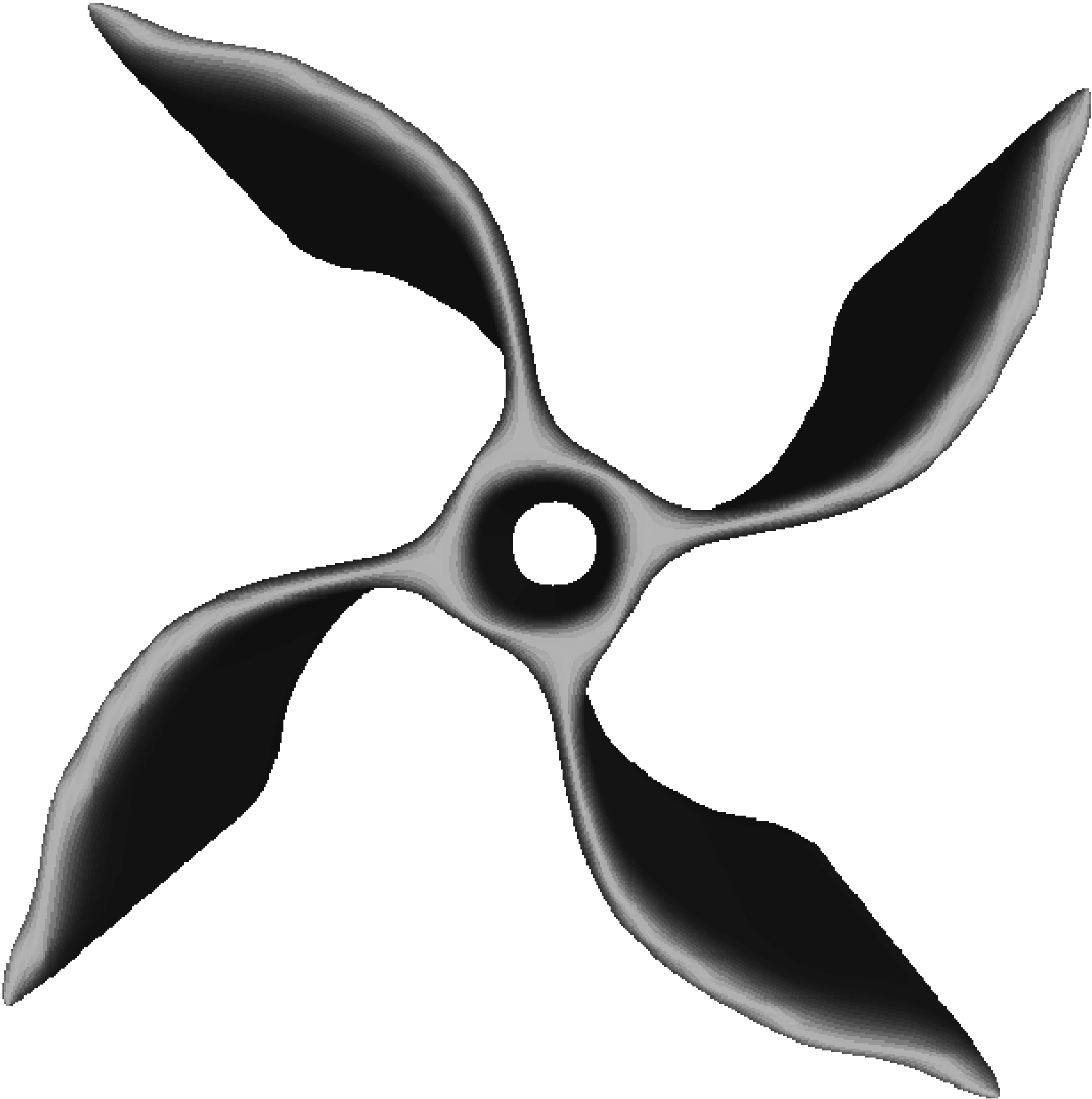} } \hspace{0.2in}
	\subfigure{\includegraphics[width=1.2in]{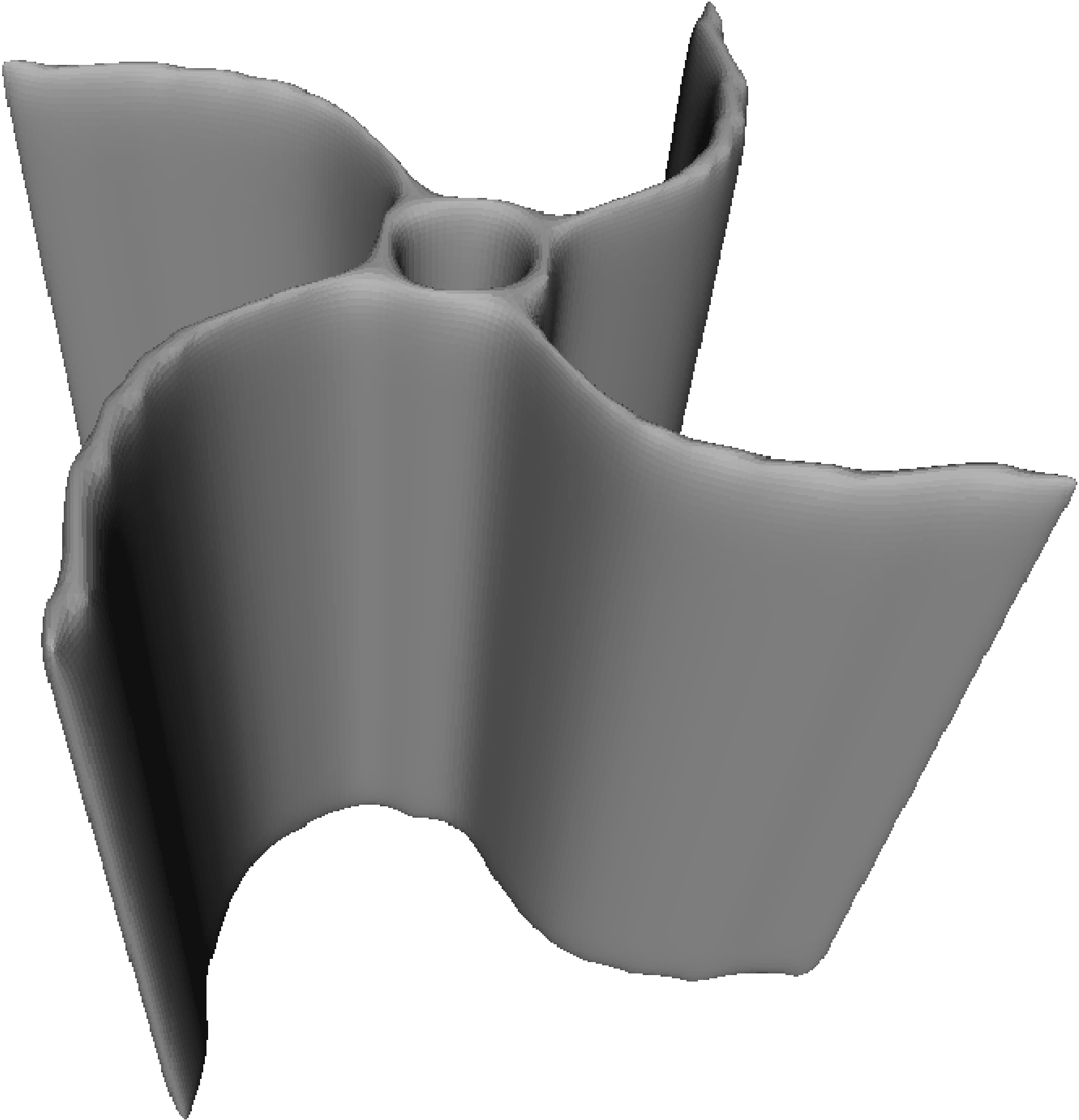} }
	\caption{Example with 50 Laplacian smoothing steps applied.}\label{fig:target-smoothed}
	\subfigure{\includegraphics[width=1.4in]{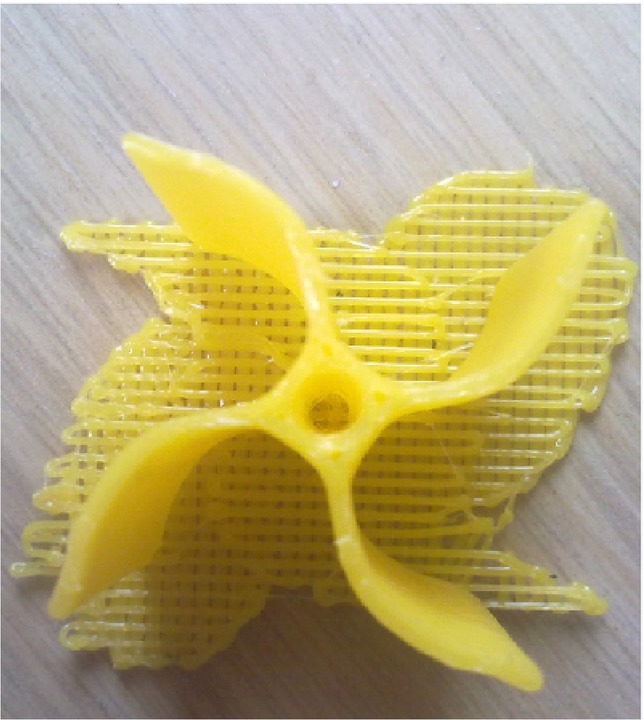} } \hspace{-0.1in}
	\subfigure{\includegraphics[width=1.4in]{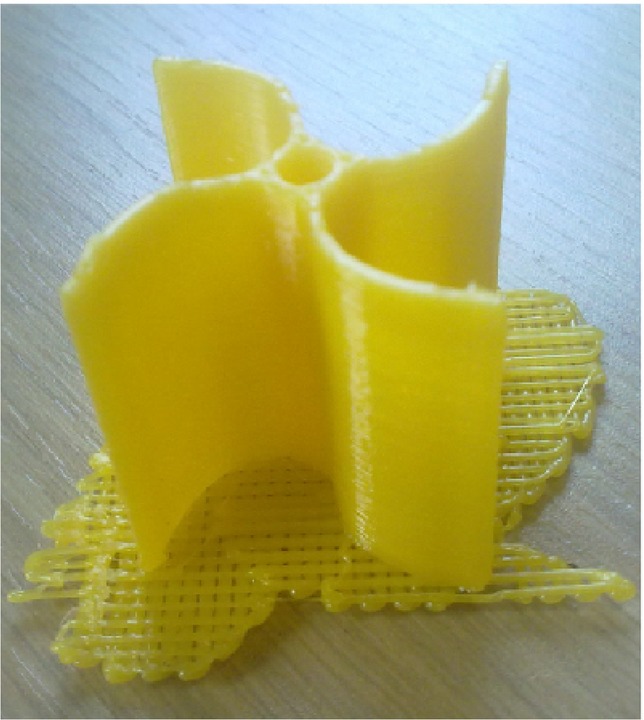} }
	\caption{Example smoothed and printed by a 3D printer; $30\times30\times30$~mm; 27~minutes printing time.}\label{fig:target-printed}
\end{figure}

An example phenotype without $z$-axis variation is shown in Figure~\ref{fig:target-unsmoothed}. When production is desired, the 3D binary voxel array is converted to stereolithography (STL) format. Once encoded in STL, it then undergoes post-processing with the application of 50 Laplacian smoothing steps using MeshLab\footnote{MeshLab is an open source, portable, and extensible system for the processing and editing of unstructured 3D triangular meshes.\ http://meshlab.sourceforge.net}; see smoothed example phenotype in Figure~\ref{fig:target-smoothed}. Finally the object is converted to printer-readable G-code and is subsequently fabricated on a BFB 3000 printer (0.25~mm resolution) using a polylactic acid (PLA) bioplastic. Figure~\ref{fig:target-printed} shows the smoothed object after fabrication.

To enable prototypes with $z$-axis variability, the genome also includes 5 additional integers in the range [-42,42], each controlling 1/6th of the $z$-axis. After drawing the initial $z$-layer as previously described, each $z$ gene transforms the $x$-$y$ genome for the next successive $z$-layer by uniformly adding the allele value, after which it is then drawn as described above. For example, with an $x$-$y$-axis genome of [2, 2, 3, 4, 5, 8, 13, 20, 34, 40] and $z$-axis genome of [2, -5, 10, 3, -2], the next $z$-layer is translated using the $x$-$y$-axis genome of [4, 4, 5, 6, 7, 10, 15, 22, 36, 42] and the following $z$-layer is translated with [1, 1, 1, 1, 2, 5, 10, 17, 31, 37], etc.

Since the interaction effects of rotor direction in the context of differing turbine morphology, relative phase angle, separation distance, positional layout, wind velocity, etc.\ are not well understood, a Boolean gene is added to designate the rotation direction of a VAWT. Thus, a total of 16 genes define an individual. In a previous CGA experiment~\citep{PreenBull:2014a} with 2 Savonius turbines positioned 0.1 diameters adjacently and in perpendicular position to the air flow, the average combined rotational speed of the final 20 counter-rotating pairs ($M=1760$, $SD=206$, $N=20$) was significantly less than the average combined rotational speed of the final 20 co-rotating pairs ($M=2048$, $SD=95$, $N=20$) using a two-tailed Mann-Whitney test ($U=17$, $p\le7.39\times10^{-7}$), showing that co-rotation was found to result in faster combined rotational speed. It can be noted that this result is similar to \cite{Sun:2012} where 2 counter-rotating S-rotors performed worse than co-rotating versions when placed at small separation distances.
 
The CGA used herein proceeds with 2 species populations, a maximum mutation step size of $\pm10$, a per allele mutation rate of 25\%, and a crossover rate of 0\%. A tournament size of 3 takes place for both selection and replacement. To help increase performance and save fabrication time, each species population initially consists of the first 10 ($>$0~rpm) individuals from a single VAWT ($z$-varying) experiment in \citep{PreenBull:2014a}, both normally rotated and counter-rotated. These individuals still retain a good degree of randomness while possessing some useful aerodynamic properties. Each species thus maintains $P=20$ individuals. The individuals in each species population are initially evaluated in collaboration with a single randomly selected individual from the other species population. Thereafter, the CGA proceeds by alternating between species after each offspring is formed and evaluated with the elite member from the other species; see algorithm outline in Algorithm~\ref{alg:cga}.

\begin{algorithm}[t]
	\SetAlgoLined%
	Generate and fabricate individuals for all species\;
	\For{each species population}{
		Select random representative from each species\;
		\For{each individual in population}{
			Evaluate\;
		}
	}
	\While{fabrication budget not exhausted}{
		\For{each species population}{
			Create an offspring using evolutionary operators\;
			Select representatives for each species\;
			Fabricate and evaluate the offspring\;
			Add offspring to species population\;
		}
	}
	\caption{Coevolutionary genetic algorithm}
	\label{alg:cga}
\end{algorithm}

For the surrogate-assisted architecture used in this paper, the basic CGA remains unchanged except that fitness evaluations are obtained from a forward pass of the genome through a neural network when the real fitness value is unknown. Initially the entire population is fabricated and evaluated on the real fitness function and added to an evaluated set. The model is trained using backpropagation for 1000 epochs; where an epoch consists of randomly selecting, without replacement, an individual from the evaluated set and updating the model weights. Each generation thereafter, the individual with the highest approximated fitness as suggested by the model and a randomly chosen unevaluated individual are fabricated and evaluated on the real fitness function and added to the evaluated set. The model weights must be reinitialised each time before training due to the temporal nature of pairing with the elite member. The genetic algorithm runs for one generation (using the model approximated fitnesses where real fitness is unknown) before the individual with the highest approximated fitness and a randomly selected unevaluated individual are evaluated with the elite member from the other species; see outline in Algorithm~\ref{alg:scga}. The model parameters, $\beta=0.3$, $\theta=0$, $elasticity=1$, $calming~rate=1$, $momentum=0$, $elasticity~rate=0$.  

\begin{algorithm}[t]
	\SetAlgoLined%
	Generate and fabricate individuals for all species\;
	\For{each species population}{
		Select random representative from each species\;
		\For{each individual in population}{
			Evaluate\;
			Add individual to species evaluated list\;
		}
	}
	\While{fabrication budget not exhausted}{
		\For{each species population}{
			Initialise model weights\;
			Train model on species evaluated list\;
			\For{each individual in population}{
				\If{individual unevaluated}{
					Set approximated fitness\;
				}
			}
			\For{population size number of times}{
				Create offspring using evolutionary operators\;
				Set offspring approximated fitness\;
				Add offspring to species population\;
			}
			Select representatives for each species\;
			Fabricate, evaluate, and add to species evaluated list, the individual with the highest approximated fitness in species\;
			Fabricate, evaluate, and add to species evaluated list, a random unevaluated individual in species\;
		}
	}
	\caption{Surrogate-assisted coevolutionary algorithm}
	\label{alg:scga}
\end{algorithm}
 
\begin{figure}[t]
	\centering 
	\includegraphics[width=4.25in]{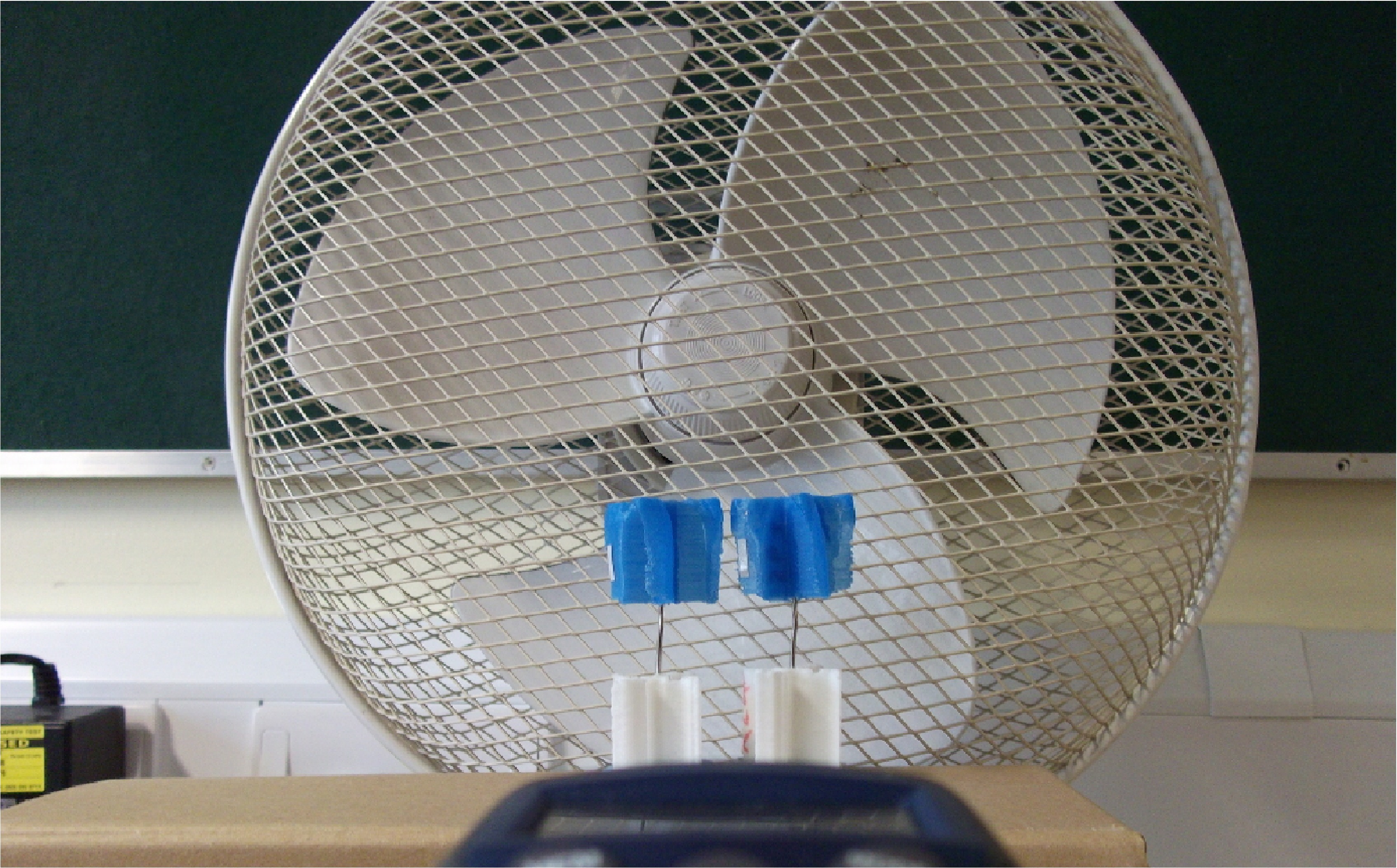}
	\caption{VAWT array experimental setup.}
	\label{fig:setup-double}
\end{figure}

The fitness of each individual is the maximum combined array rotational speed achieved over the period of 1~minute during the application of constant wind generated by a propeller fan after fabrication by a 3D printer. The rotational speed is the significant measure of aerodynamic efficiency since the design space is constrained (including rotor radius and turbine height). However, in future work, the power generated will be preferred, which will take into account any slight weight variations that may affect performance. The rotational speed is here measured in number of revolutions per minute (rpm) using a digital photo laser tachometer (PCE-DT62; PCE Instruments UK Ltd) by placing a $10\times2$~mm strip of reflecting tape on the outer tip of one of the individual's blades. The experimental setup can be seen in Figure~\ref{fig:setup-double}, which shows the $30$~W, 3500~rpm, 304.8~mm propeller fan, which generates 4.4~m/s wind speed, and 2 turbines mounted on rigid metal pins 1~mm in diameter and positioned 33~mm adjacently and 30~mm from the propeller fan. That is, there is a 3~mm spacing between the blades at their closest point.

\section{Experiments}

\subsection{Modelling Techniques}

To explore whether there is any significant advantage in replacing the neural network used previously as the surrogate model, several algorithms using the Weka 3.6.10 machine learning collection\footnote{Weka is an open source data mining algorithm collection. http://www.cs.waikato.ac.nz/ml/weka/} were run over the data generated from \citep{PreenBull:2014a}. It is important to note that a surrogate model with a larger fitness prediction error may be more accurate in predicting the rank order of individuals than a model with lower error. Since the rank order is the guiding influence on the evolutionary process, alternative quality measures for approximate models have therefore been suggested. However, \cite{Bischl:2012} found no difference between using the root mean squared error (RMSE), mean absolute error (MAE), or Spearman's rho, which measures the correlation between the actual and predicted ranking. Furthermore, in practice, optimisation based on the fitness accuracy has often been found to perform the best~\citep[e.g.,][]{Husken:2005}. Thus, the MAEs of the fitness predictions are here used as the performance metric.

Table~\ref{table:mae} shows the average MAEs for various algorithms predicting the fitness of evolved individuals. Each number represents the average MAE over 100 runs using 10-fold cross-validation (CV). The result is marked in bold font where it is statistically different from the standard MLP at the 95\% confidence level using a two-sample $t$-test assuming unequal variances. GA is the $z$-axis varying GA-only single VAWT experiment ($N=100$); SGA is the surrogate-assisted GA $z$-axis varying VAWT experiment ($N=100$); CGA-1 is the first species from the CGA experiment ($N=80$); CGA-2 is the second species from the CGA experiment ($N=80$); SCGA-1 is the first species from the surrogate-assisted CGA experiment ($N=80$); and SCGA-2 is the second species from the surrogate-assisted CGA experiment ($N=80$). Default Weka parameters were used unless otherwise specified. 

The results show that there is little statistical difference between the algorithms in predicting the fitness of the evolved individuals. This is perhaps due to the very small and noisy training samples available for this task. In the rest of this paper we therefore continue with neural network surrogate models, which are well suited to problems with a high-dimensional input space (design space) and limited number of samples~\citep{Jin:2005}. In addition, neural network modelling frequently outperforms more complex approaches such as coevolutionary active learning~\citep{LyLipson:2014}.

\begin{table}[t]
	\caption{Average MAEs (100 runs using 10-fold CV) for various algorithms predicting the fitness of evolved individuals from the experiments in \citep{PreenBull:2014a}. The result is marked in bold font where it is statistically different from the standard MLP at the 95\% confidence level using a two-sample $t$-test assuming unequal variances.}
    \centering
    \begin{tabular}{l r r r r r r }
        \hline
		Algorithm & GA & SGA & CGA-1 & CGA-2 & SCGA-1 & SCGA-2\\
        \hline
		Standard MLP & 197.89 & 237.86 & 430.30 & 432.64 & 462.57 & 394.04 \\
		Linear Regression & 158.91 & {\bf 170.78} & {\bf 275.13} & 390.79 & 361.97 & 379.08 \\
		Isotonic Regression & 135.57 & {\bf 145.97} & 327.22 & 462.68 & 367.10 & {\bf 523.73} \\
		Pace Regression  & 160.75 & 172.57 & {\bf 257.57} & 397.99 & 380.85 & 384.50 \\
		Least Med Squares  & 208.80 & 237.78 & {\bf 306.71} & 432.26 & 334.00 & 447.14 \\
		SMO Regression  & 152.70 & 191.33 & {\bf 280.94} & 370.18 & {\bf 334.65} & 356.75 \\
		MLP $N=1000$ $H=10$  & 207.21 & 241.89 & 451.91 & 427.81 & 503.93 & 418.42 \\
		MLP $N=1000$ $H=100$ & 265.10 & 202.33 & 385.61 & 414.38 & 416.14 & 408.88 \\
		MLP $N=10000$ $H=10$ & 256.52 & 305.41 & 562.50 & 481.27 & {\bf 654.00} & 487.14 \\
		MLP $N=500$ $H=10$ & 191.63 & 216.88 & 397.40 & 401.16 & 405.01 & 401.27 \\
		MLP $N=500$ $H=1$ & 142.13 & 174.45 & 384.69 & 338.37 & 410.49 & 333.47 \\
		MLP $N=1000$ $H=1$ & 142.32 & 177.98 & 395.50 & 340.86 & 418.94 & 343.24 \\
		MLP $N=10000$ $H=1$ & 159.20 & 193.07 & 429.34 & 364.05 & 446.37 & 366.36 \\
		MLP $N=500$ $H=2$ & 154.28 & 181.66 & 392.26 & 389.26 & 390.21 & 337.27 \\
		MLP $N=1000$ $H=2$ & 156.30 & 181.51 & 408.23 & 407.67 & 408.15 & 358.50 \\
		Gaussian Processes & 168.00 & 193.03 & {\bf 281.32} & 369.74 & {\bf 323.65} & 361.45 \\
		RBF $B=2$ & 172.28 & 180.44 & 432.09 & 533.81 & 482.91 & {\bf 596.46} \\
		RBF $B=10$ & 213.94 & {\bf 158.23} & 1253.08 & 512.91 & 404.72 & 3981.20 \\
		REP Tree& 196.11 & {\bf 164.33} & 346.77 & 343.68 & 394.93 & 377.64 \\
		Decision Stump & 169.99 & 193.78 & 325.33 & 488.70 & 390.42 & {\bf 519.27} \\
		M5P & 165.70 & {\bf 168.34} & {\bf 311.37} & 337.33 & 355.37 & 350.04 \\
        \hline
    \end{tabular}
    \label{table:mae}
\end{table}

\subsection{Windowing}

In interactive evolutionary computation, a user's evaluation is relative within each generation and therefore fitness values from generations long past may be different to recent generations even if the evaluated candidate is identical. Consequently, \cite{WangTakagi:2005} found that a neural network surrogate model using only recent training data was more effective than using all past data. In addition, many other surrogate-assisted approaches have used fixed length training sets~\cite[e.g.,][]{Bull:1997a}, although did so to save computational time rather than intentionally aid convergence. Due to the temporal nature of partnering with the elite species members it may be advantageous to exclude older training samples for SCGAs. To explore whether there is any benefit from using different windowing approaches to training the surrogate model, 13 different models from Weka were run across the 4 datasets consisting of the 80 evaluated individuals in each species from the coevolution experiments; 2 species for CGA and 2 for SCGA. The final 2 individuals from each set were used for testing and were excluded from any training.

The models used are as follows and Weka default parameters were used unless otherwise stated: (1) Gaussian Processes (Support Vector, RBF Kernel); (2) Isotonic Regression; (3) Linear Regression (M5 attribute selection); (4) Multilayer Perceptron; (5) SMOreg (RBF Kernel, RegSMO Improved); (6) RBF Network ($B=2$); (7) RBF Network ($B=100$); (8) Multilayer Perceptron ($N=1000$, $H=10$, $M=0$); (9) Multilayer Perceptron ($N=500$, $H=1$); (10) Multilayer Perceptron ($N=1000$, $H=1$); (11) Multilayer Perceptron ($N=500$, $H=5$); (12) Multilayer Perceptron ($N=1000$, $H=5$); (13) Multilayer Perceptron ($N=1000$, $H=100$). 

One set of experiments used the full 78 training samples for each dataset ($T=78$); one set used the most recent 40 training samples ($T=40$); one set used the most recent 20 training samples ($T=20$); and one set used the most recent 10 training samples ($T=10$). The average of the 13 algorithm's MAEs over the 4 datasets with $T=20$ ($M=197.11$, $SD=30.54$, $N=13$) is significantly less than $T=78$ ($M=336.39$, $SD=18.34$, $N=13$) using a two-sample $t$-test assuming unequal variances, $t(14)=4.76$, $p=.0003$. In addition, the average MAE of $T=20$ is also significantly less than $T=40$ ($M=269.94$, $SD=16.43$, $N=13$), $t(16)=3.31$, $p=.0045$. However, the average MAE of $T=10$ ($M=190.14$, $SD=15.03$, $N=13$) was not significantly less than $T=20$, $t(17)=0.74$, $p=.47$.

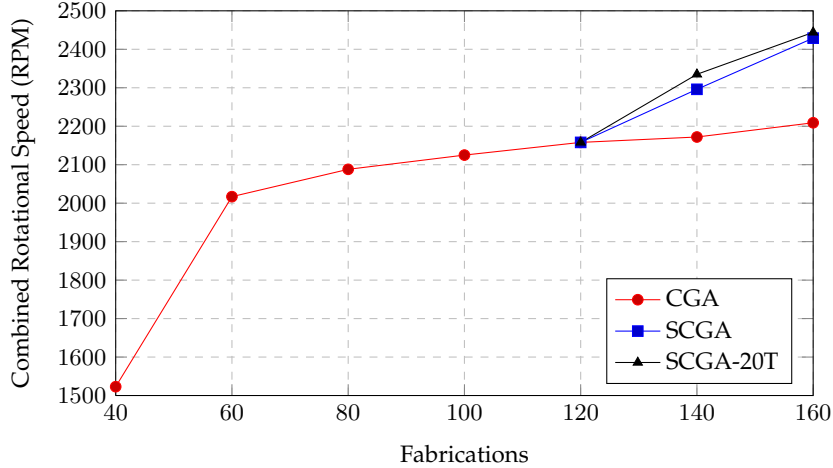
\begin{figure}[t]
	\centering 
	\begin{tikzpicture}[font=\figfontsize] 
		\begin{axis}[
				width=\graphwidth,
				height=\graphheight,
				/pgf/number format/.cd,
				1000 sep={},
				xlabel=Fabrications,
				ylabel=Combined Rotational Speed (RPM),
				grid=major,
				grid style={dashed, gray!50},
				ymin=1500,
				ymax=2500,
				ytick={1500,1600,...,2500},
				xmin=40,
				xmax=160,
				legend entries={CGA,SCGA,SCGA-20T},
				legend style={nodes=right},
				legend pos= south east,
				cycle list name=mark list,
			]
			\addplot+[red] table [x=a, y=b, col sep=comma] {coevo-evals.dat};
			\addplot+[blue] table [x=a, y=c, col sep=comma] {coevo-evals.dat};
			\addplot+[black] table [x=a, y=d, col sep=comma] {coevo-evals.dat};
		\end{axis}
	\end{tikzpicture}
	\caption{Array rotational speed-based evolution. Fittest array pairs. CGA (circle), SCGA (square), and SCGA-20T (triangle). The SCGAs are used for comparison only after 120 evaluations (i.e.,\ 3 generations) of the CGA since sufficient training data is required for the surrogate models.}
	\label{fig:res-t20}
\end{figure}

\begin{figure}[t]
	\centering 
	\subfigure[CGA; 2209~rpm.]{ 
		\includegraphics[width=\figwidth]{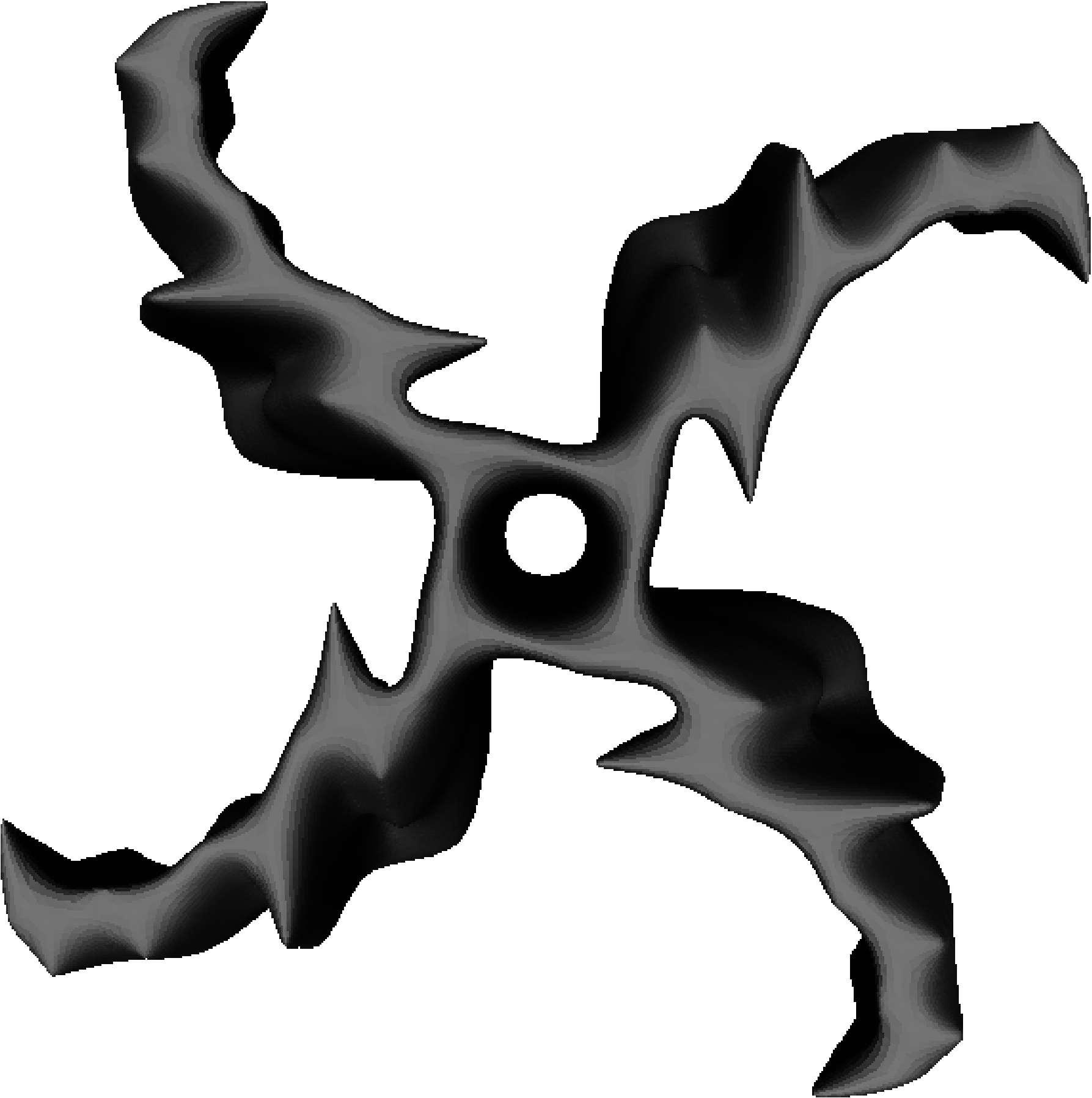} \includegraphics[width=\figwidth]{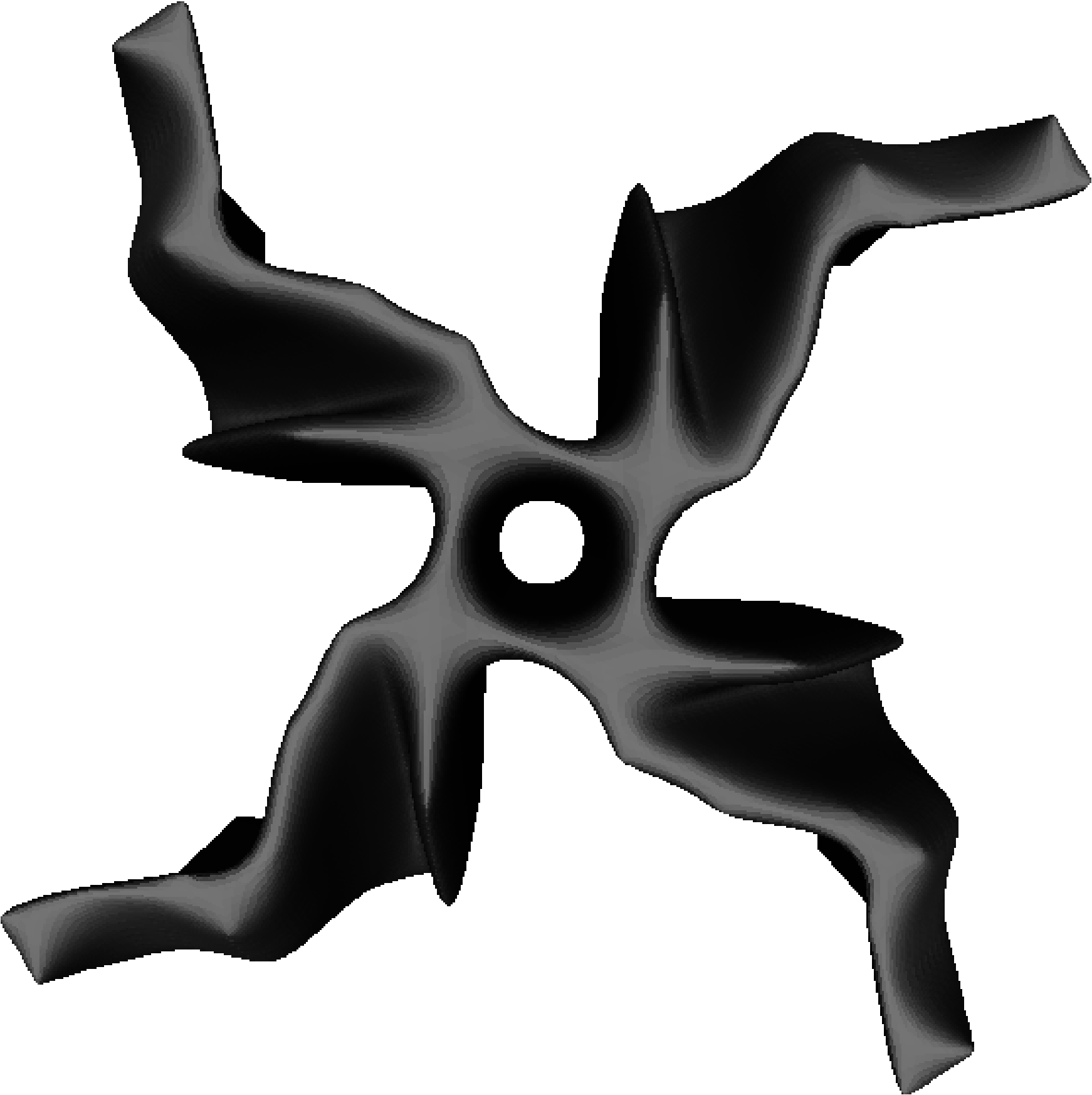} \hspace{0.3in}
		\includegraphics[width=\figwidth]{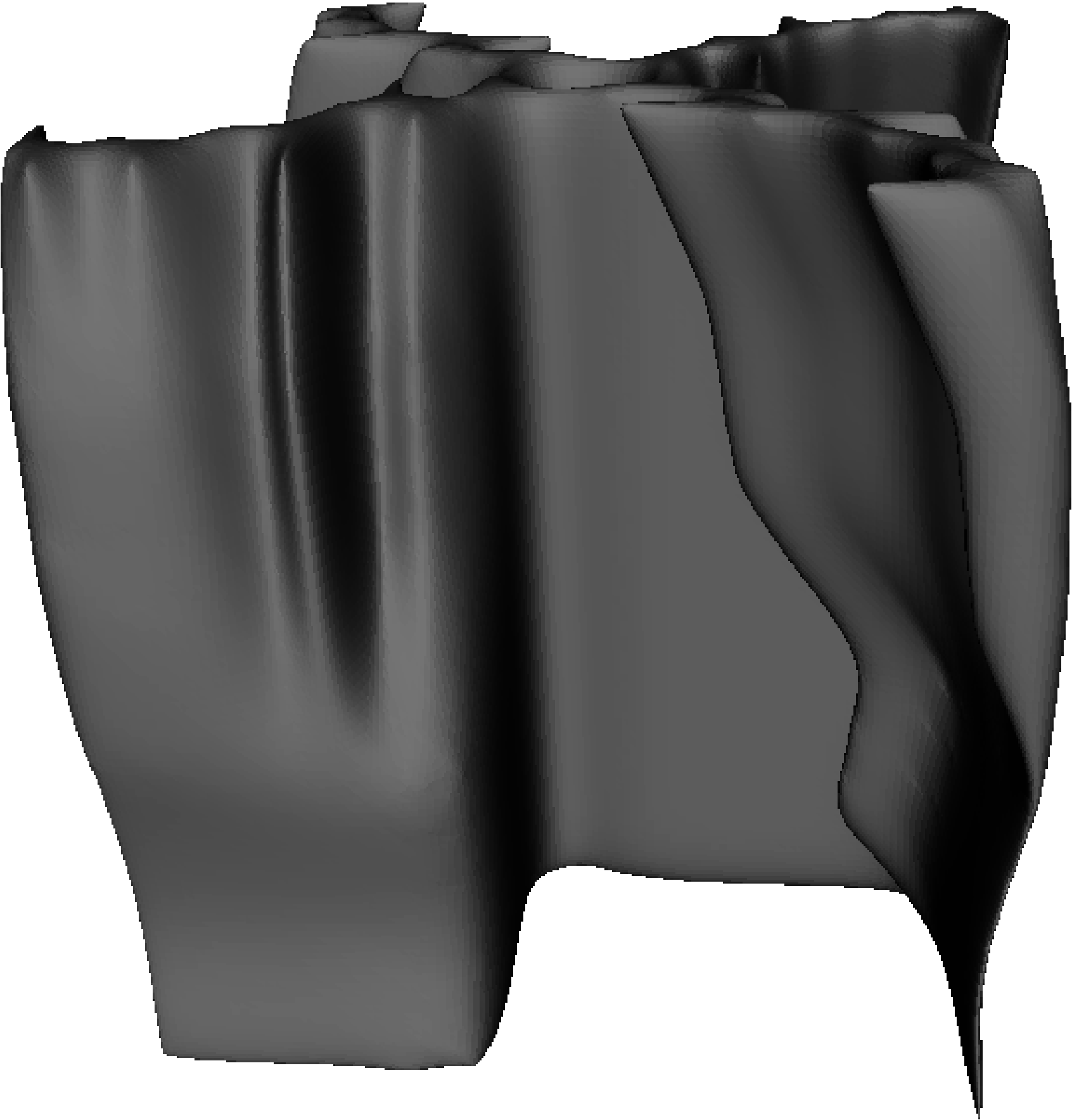} \includegraphics[width=\figwidth]{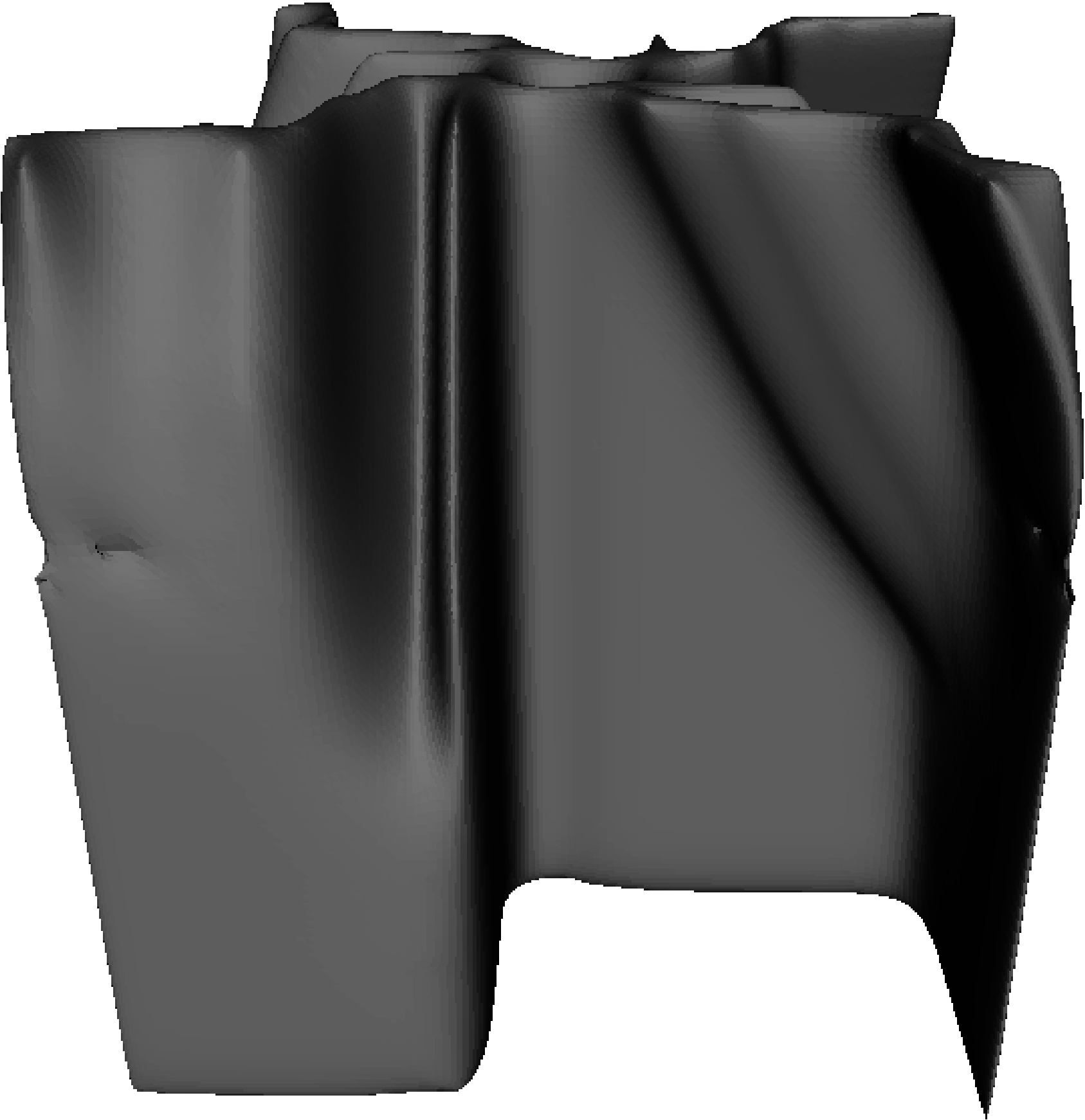} } \\
	\subfigure[SCGA; 2429~rpm.]{ 
		\includegraphics[width=\figwidth]{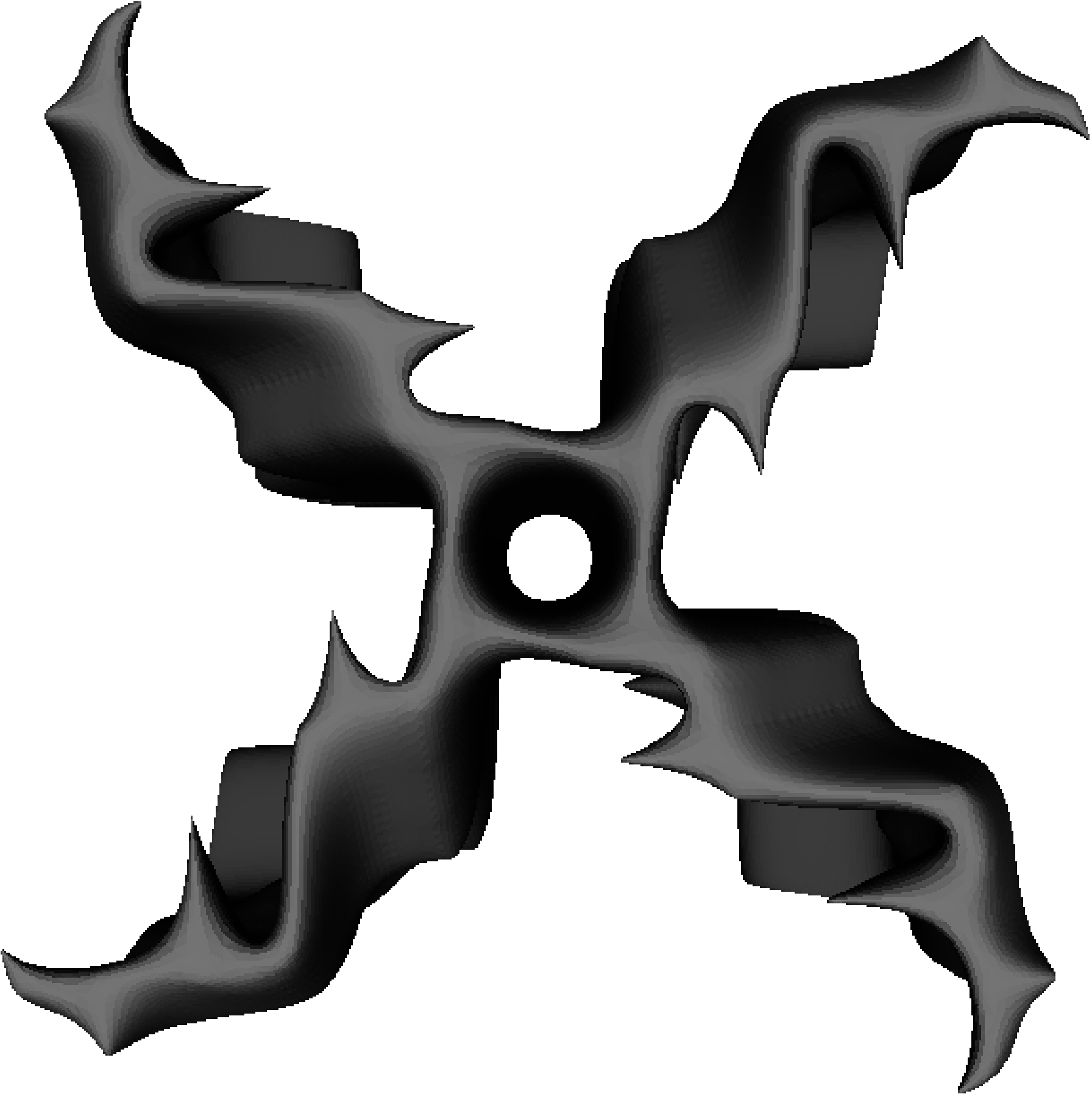} \includegraphics[width=\figwidth]{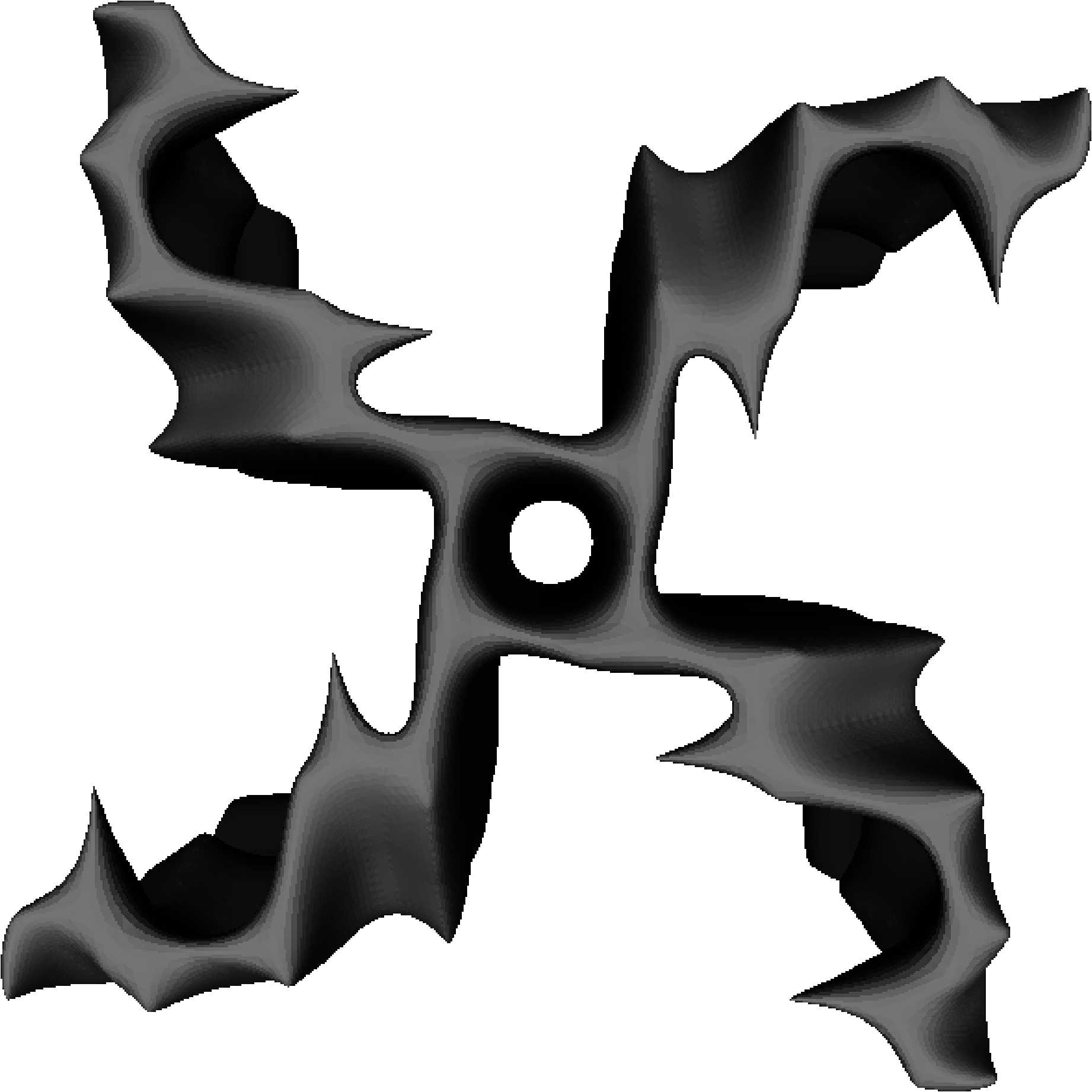} \hspace{0.3in}
		\includegraphics[width=\figwidth]{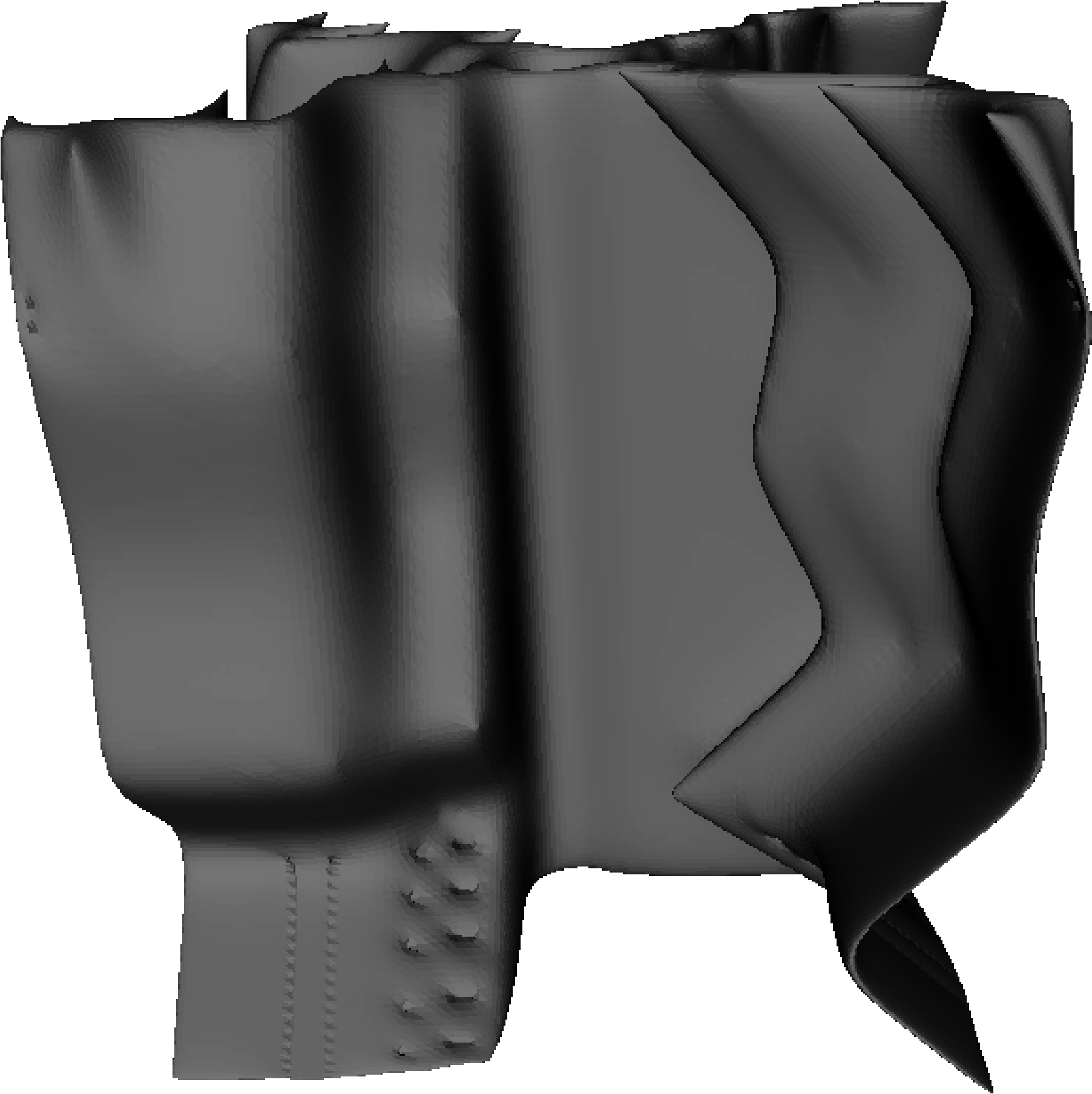} \includegraphics[width=\figwidth]{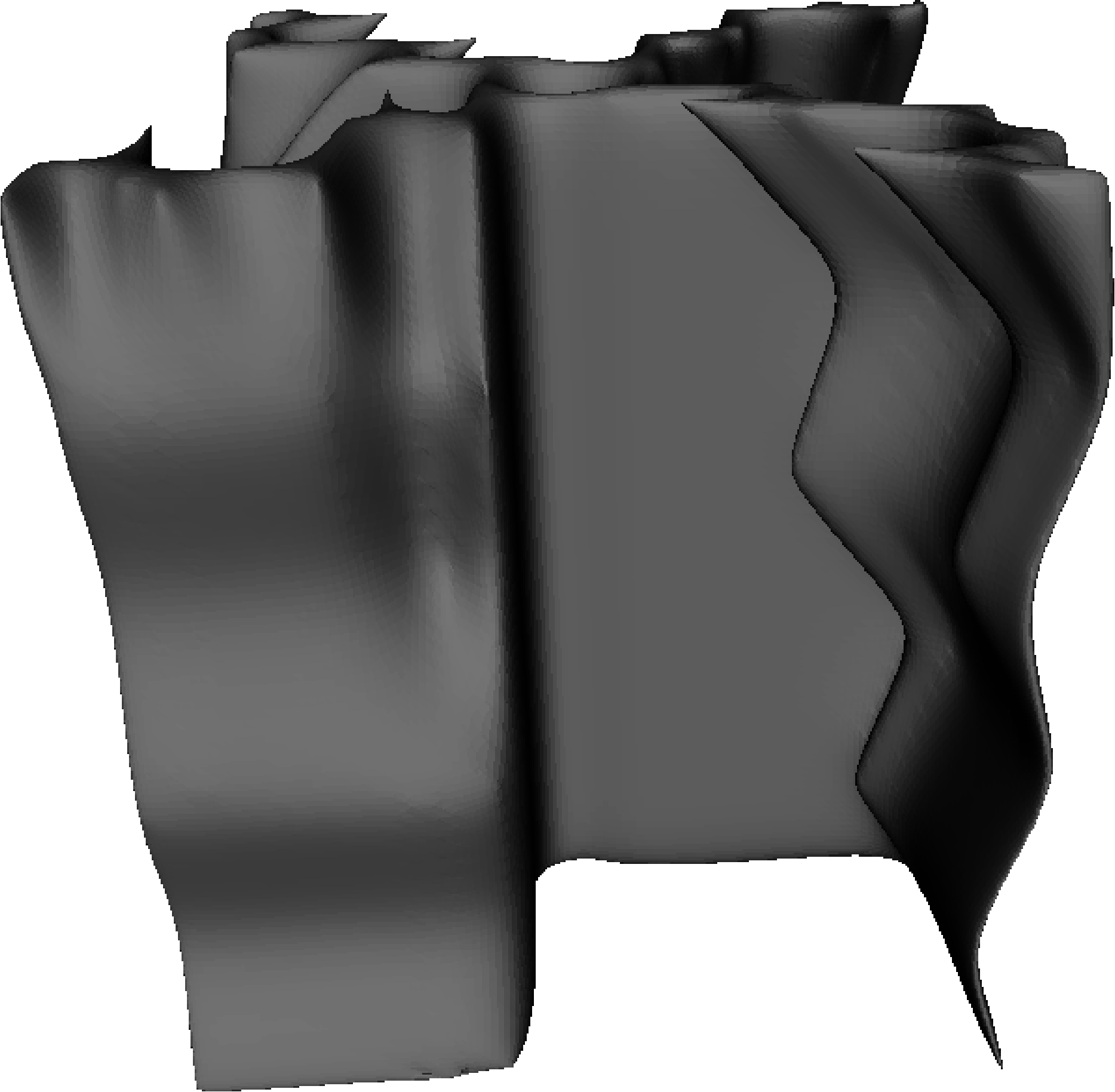} } \\
	\subfigure[SCGA-20T; 2444~rpm.]{ 
		\includegraphics[width=\figwidth]{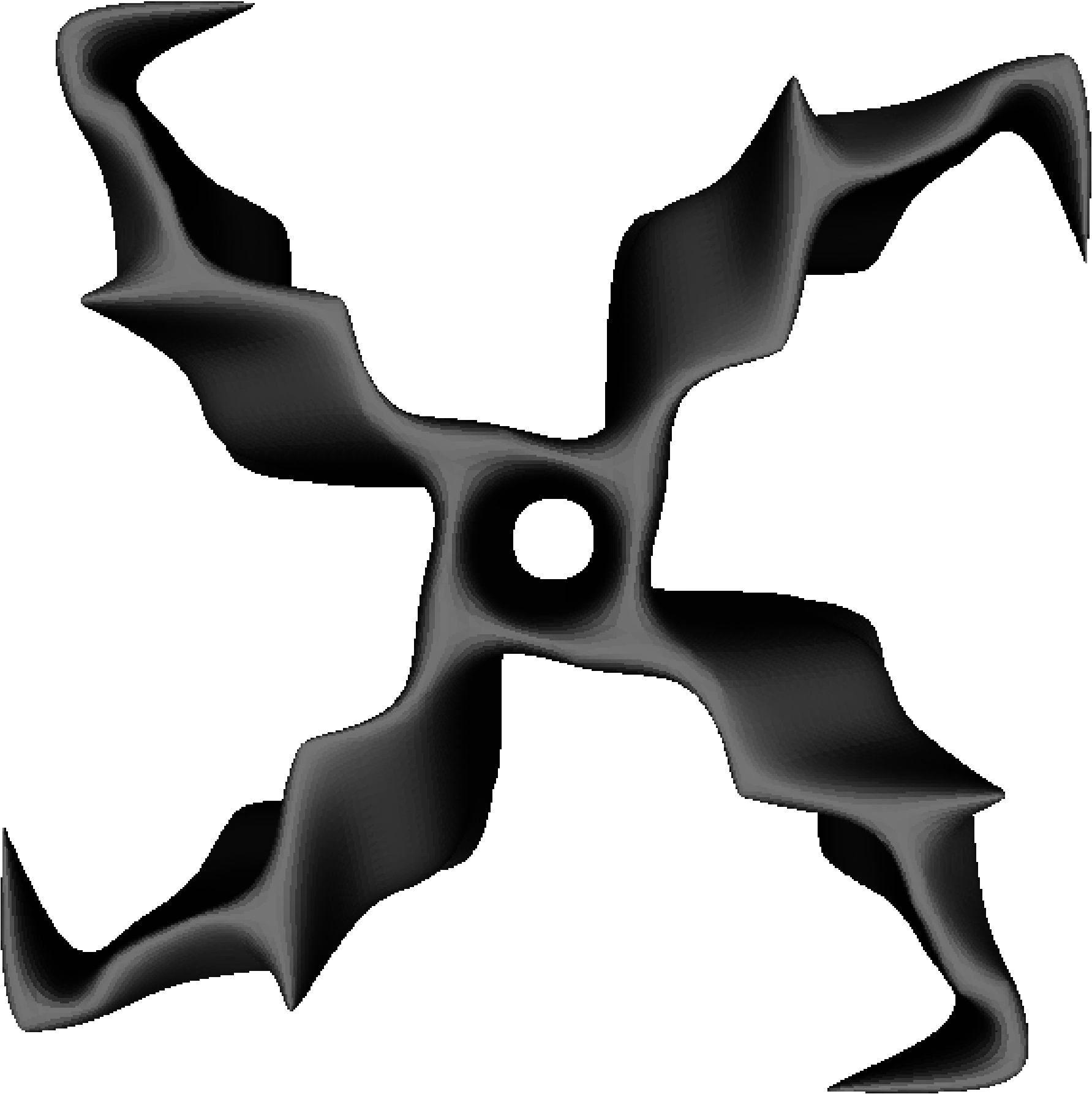} \includegraphics[width=\figwidth]{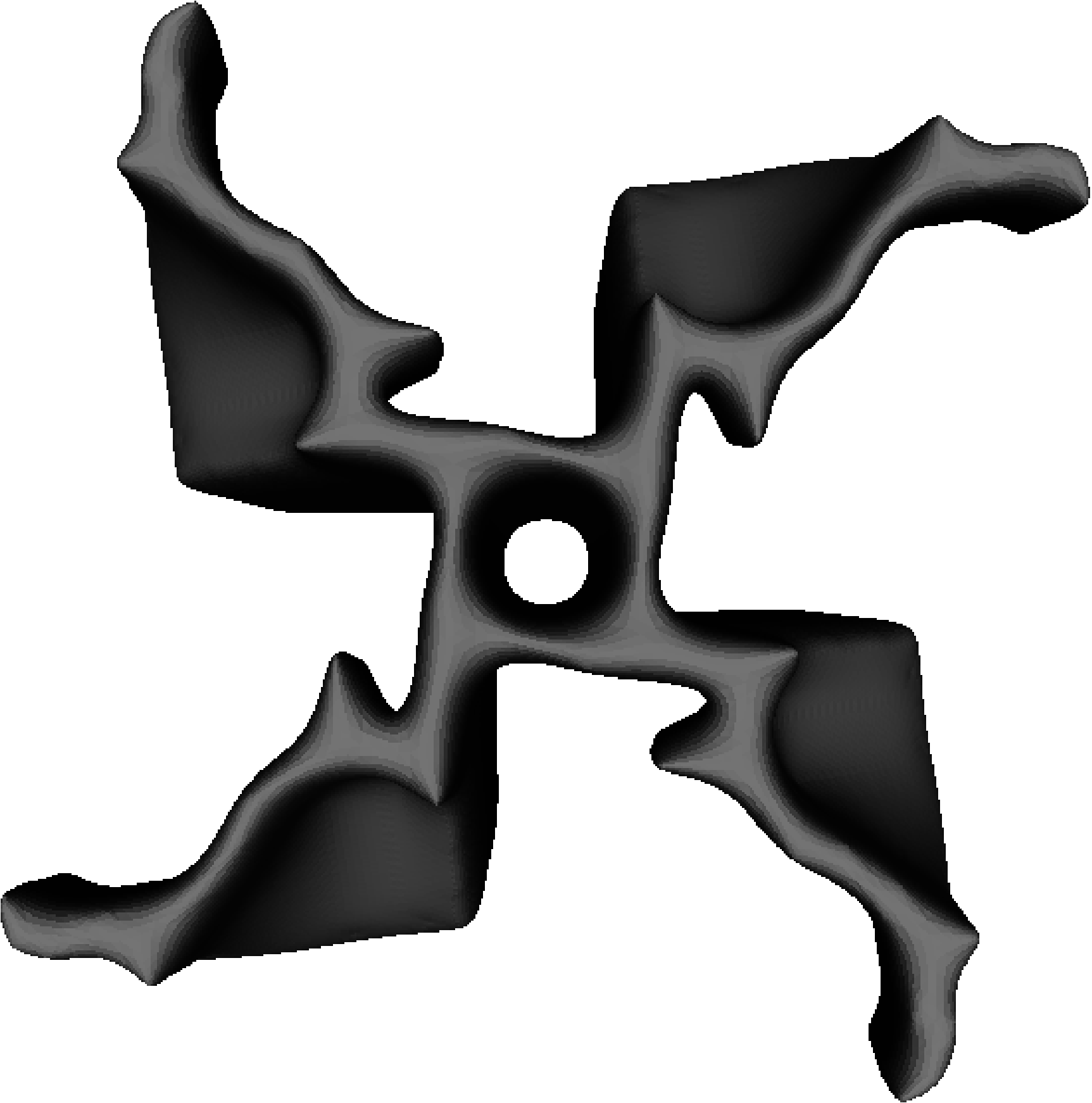} \hspace{0.3in}
		\includegraphics[width=\figwidth]{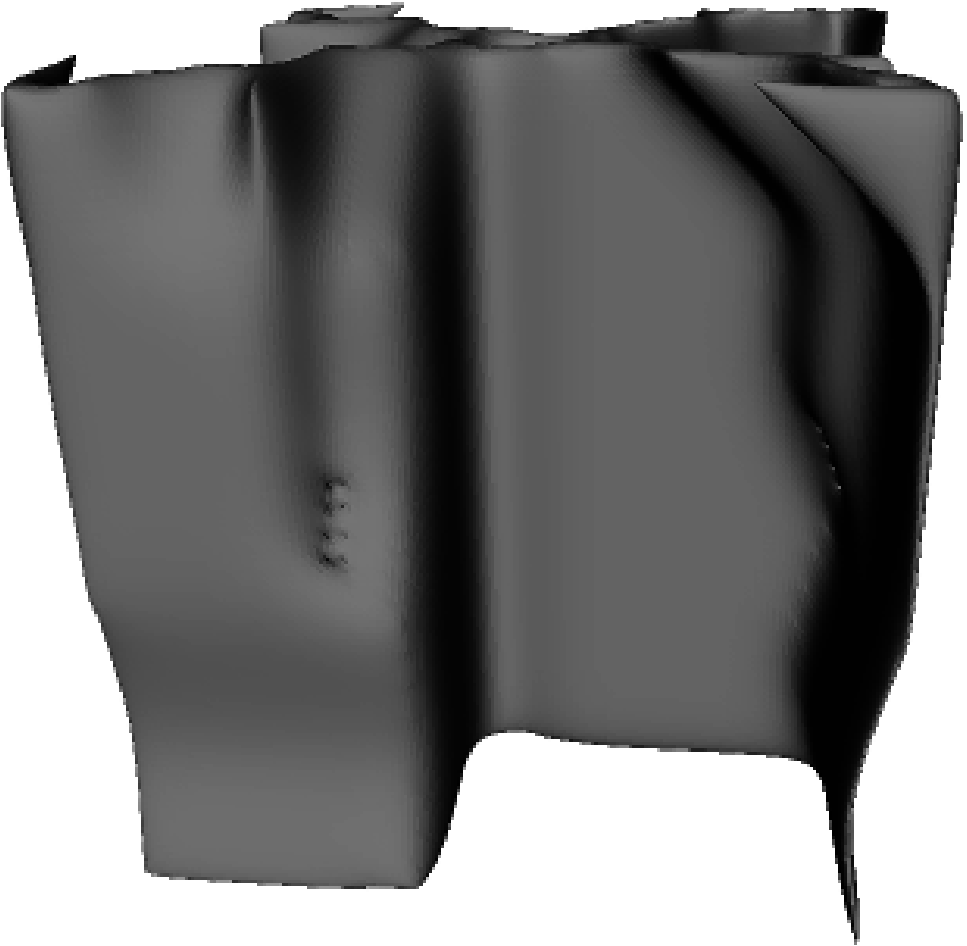} \includegraphics[width=\figwidth]{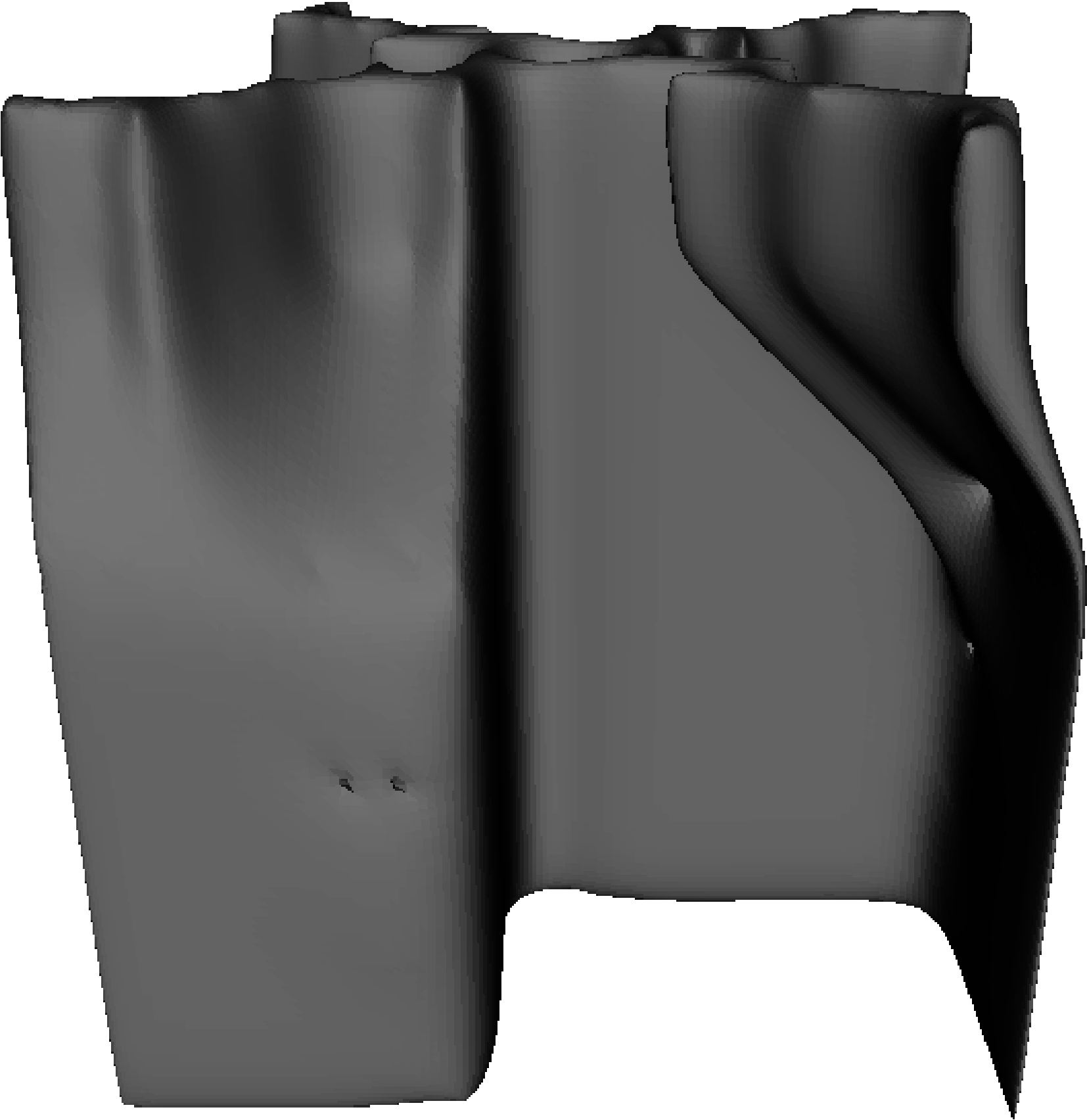} }
	\caption{Fittest evolved interacting VAWTs after 160 fabrications.}
	\label{fig:best-evolved}
\end{figure}

These results suggest that using the most recent 20 training samples (i.e., $P$) produces a more accurate model than using the full evaluation set. The SCGA was therefore rerun, as before however using only the most recent 20 evaluated individuals for training (SCGA-20T), and the results are shown in Figure~\ref{fig:res-t20}. The average rotational speed of the original SCGA ($M=2112$, $SD=307$, $N=40$) was significantly greater than CGA ($M=1905$, $SD=223$, $N=40$) using a two-tailed Mann-Whitney test, $U=331$, $p\le6.4\times10^{-6}$. Furthermore, the fittest array combination designed by SCGA (2429~rpm) was greater than CGA (2209~rpm) after 160 fabrications. However, the average rotational speed of SCGA-20T ($M=1971$, $SD=386$, $N=40$) is not significantly different to SCGA ($M=2112$, $SD=307$, $N=40$) using a two-tailed Mann-Whitney test, $U=629$, $p\le.1$, and the fittest array combination was nearly identical after 160 fabrications; SCGA-20T 2444~rpm vs. SCGA 2429~rpm. The fittest CGA, SCGA and SCGA-20T evolved arrays after 160 fabrications can be seen in Figure~\ref{fig:best-evolved}.

\subsection{Model Enhanced Local Search}

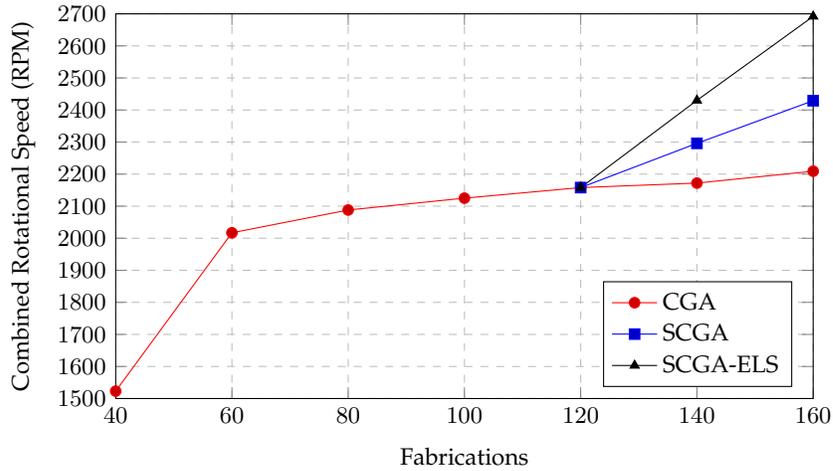
\begin{figure}[t]
	\centering 
	\begin{tikzpicture}[font=\figfontsize] 
		\begin{axis}[
				width=\graphwidth,
				height=\graphheight,
				/pgf/number format/.cd,
				1000 sep={},
				xlabel=Fabrications,
				ylabel=Combined Rotational Speed (RPM),
				grid=major,
				grid style={dashed, gray!50},
				ymin=1500,
				ymax=2700,
				ytick={1500,1600,...,2700},
				xmin=40,
				xmax=160,
				legend entries={CGA,SCGA,SCGA-ELS},
				legend style={nodes=right},
				legend pos= south east,
				cycle list name=mark list,
			]
			\addplot+[red] table [x=a, y=b, col sep=comma] {coevo-evals.dat};
			\addplot+[blue] table [x=a, y=c, col sep=comma] {coevo-evals.dat};
			\addplot+[black] table [x=a, y=e, col sep=comma] {coevo-evals.dat};
		\end{axis}
	\end{tikzpicture}
	\caption{Array rotational speed-based evolution. Fittest array pairs. CGA (circle), SCGA (square), and SCGA-ELS (triangle). The SCGAs are used for comparison only after 120 evaluations (i.e.,\ 3 generations) of the CGA since sufficient training data is required for the surrogate models.}
	\label{fig:res-pre}
\end{figure}

Typically $n$ individuals are evaluated with the real fitness function each generation, where $n<P$. This results in a population consisting of a mix of individuals evaluated with the real fitness function and those whose fitnesses are approximated. However, \cite{Ulmer:2004} introduced a pre-selection approach to the surrogate assistance of ESs. The approach consists of creating $\lambda_{Pre}$ offspring, where $\lambda_{Pre}>\lambda$, and evaluating each with the surrogate model; whereupon the $\lambda$ individuals with the highest approximated fitness are then selected to form the offspring population and are evaluated with the real fitness function. The main difference is that with pre-selection all offspring are generated from parent individuals evaluated directly with the real fitness function. This enables an enhanced local search through the evaluation of a large number of offspring while preventing the evolutionary search from drifting too far from the evaluated design space, which can occur when repeatedly creating offspring from approximated individuals. They found that the bigger $\lambda_{Pre}$ is, the better the algorithm performed. Here we explore the use of a SCGA with an enhanced local search, SCGA-ELS. The CGA runs as normal except that each time a parent is chosen, $m$ number of offspring are created and evaluated with the surrogate model, and the single offspring with the highest approximated fitness is then fabricated and evaluated with the real fitness function before being added to the population. See outline in Algorithm~\ref{alg:scga-pre}. Here $m=1000$. The results are shown in Figure~\ref{fig:res-pre}. 

\begin{algorithm}[t]
	\SetAlgoLined%
	Generate and fabricate individuals for all species\;
	\For{each species population}{
		Select random representative from each species\;
		\For{each individual in population}{
			Evaluate\;
			Add individual to species evaluated list\;
		}
	}
	\While{fabrication budget not exhausted}{
		\For{each species population}{
			Initialise model weights\;
			Train model on species evaluated list\;
			Select a parent using tournament selection\;
			\For{$m$ number of times}{
				Create an offspring using evolutionary operators\;
				Set offspring approximated fitness\;
			}
			Select the single offspring with the highest approximated fitness\;
			Select representatives for each species\;
			Fabricate and evaluate the selected offspring\;
			Add the selected offspring to the species population\;
			Add the selected offspring to the species evaluated list\;
		}
	}
	\caption{SCGA with enhanced local search}
	\label{alg:scga-pre}
\end{algorithm}
 
\begin{figure}[t]
	\centering 
	\includegraphics[width=\figwidth]{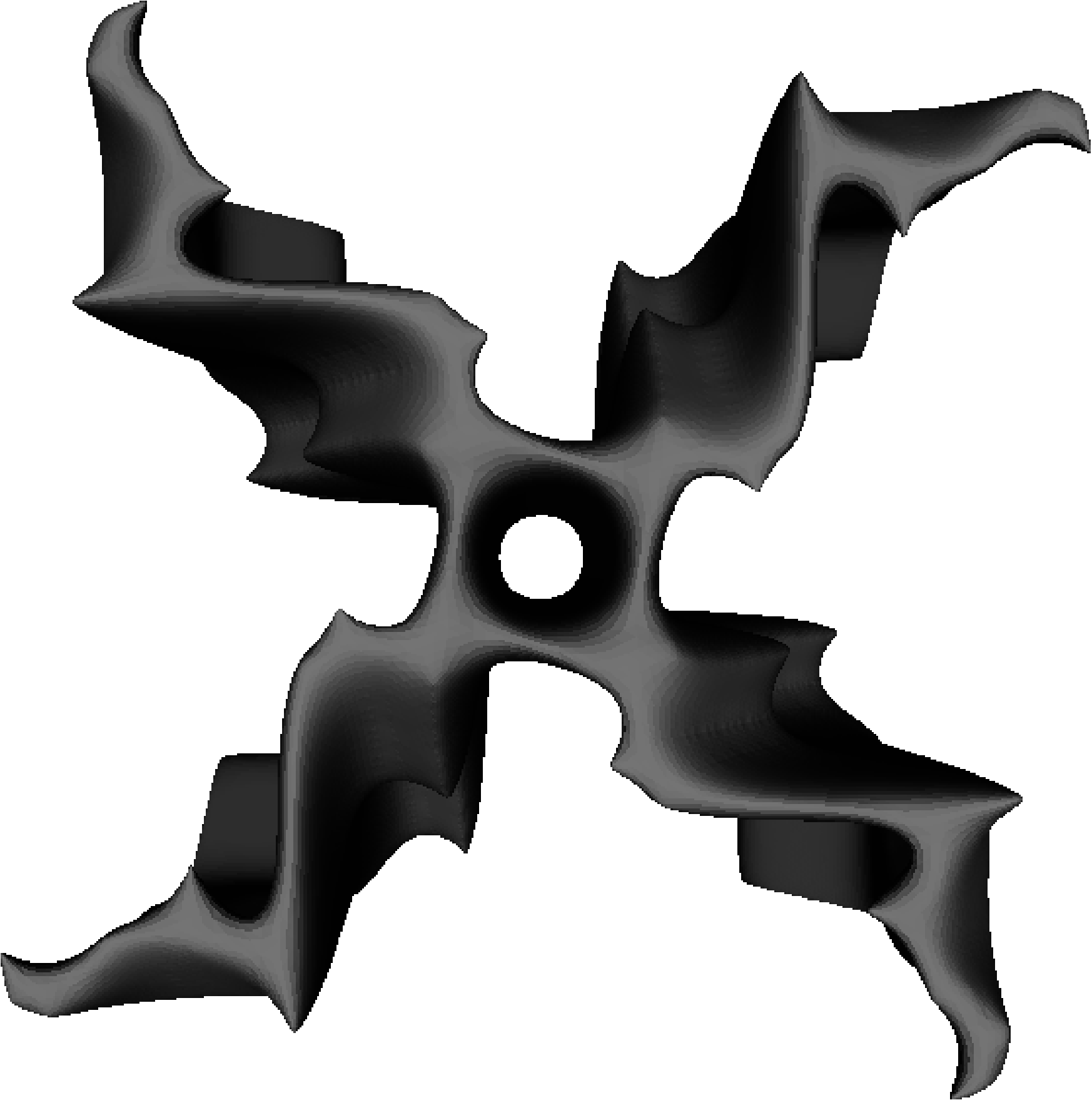} \includegraphics[width=\figwidth]{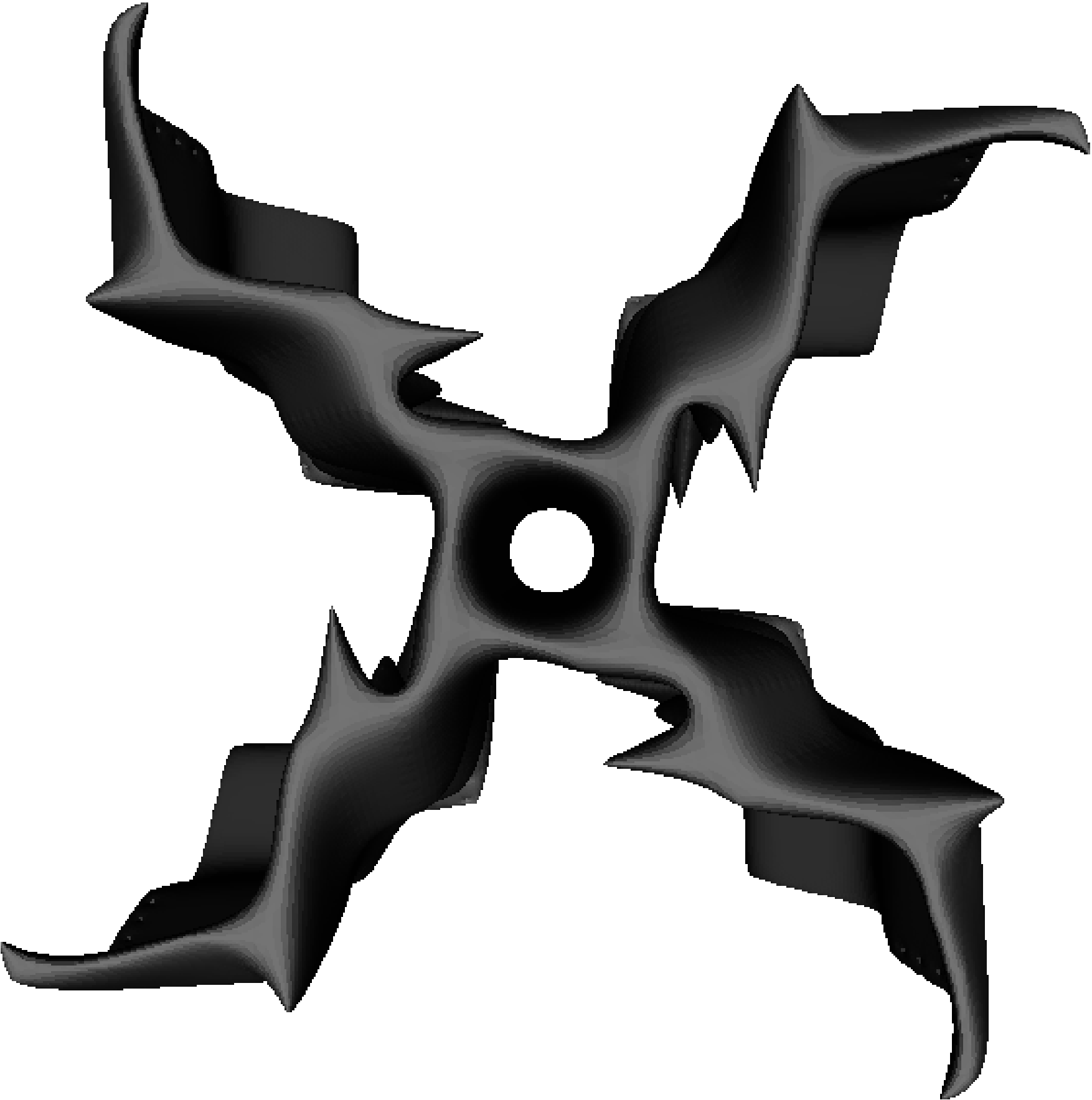} \hspace{0.3in}   
	\includegraphics[width=\figwidth]{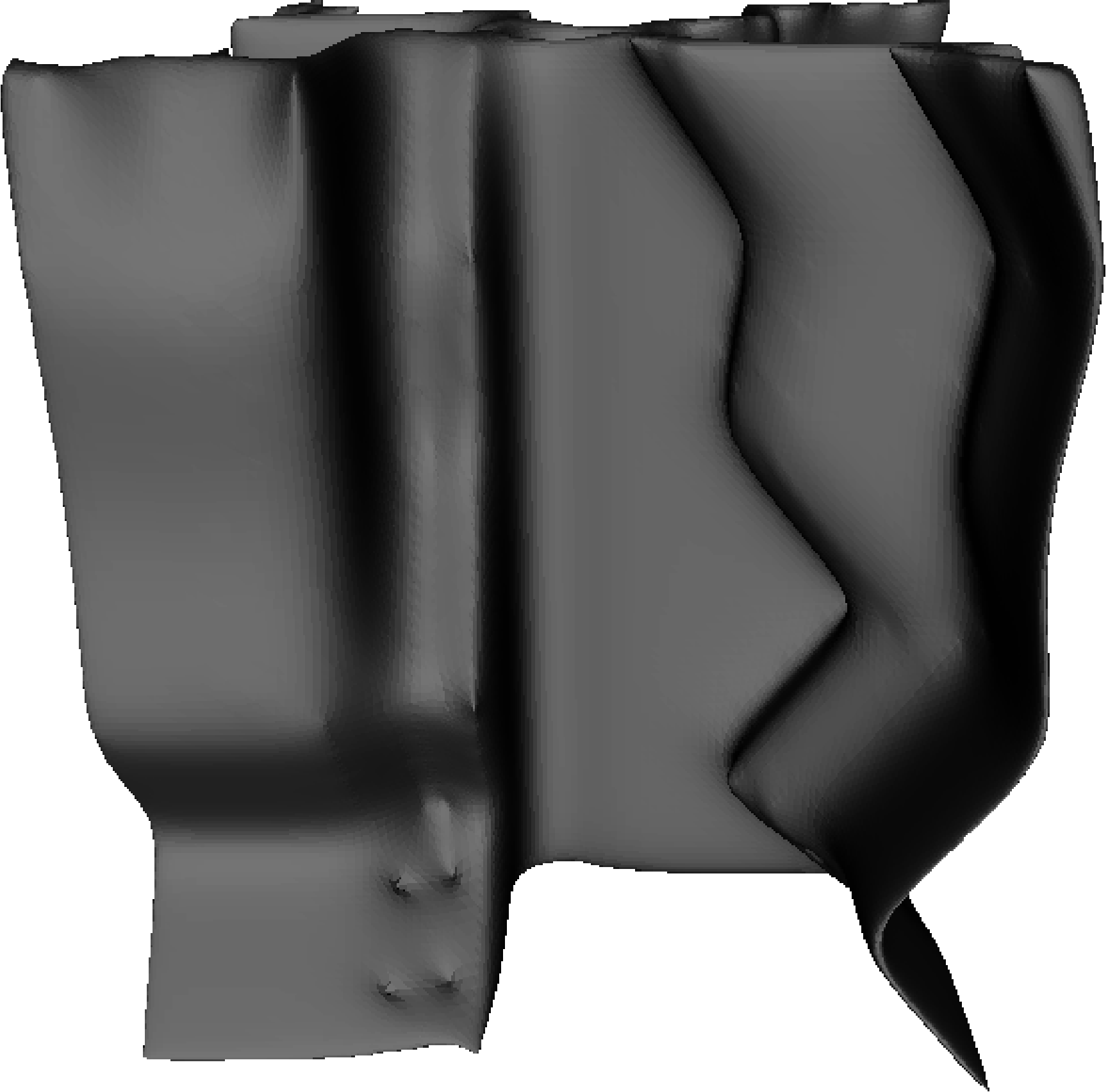} \includegraphics[width=\figwidth]{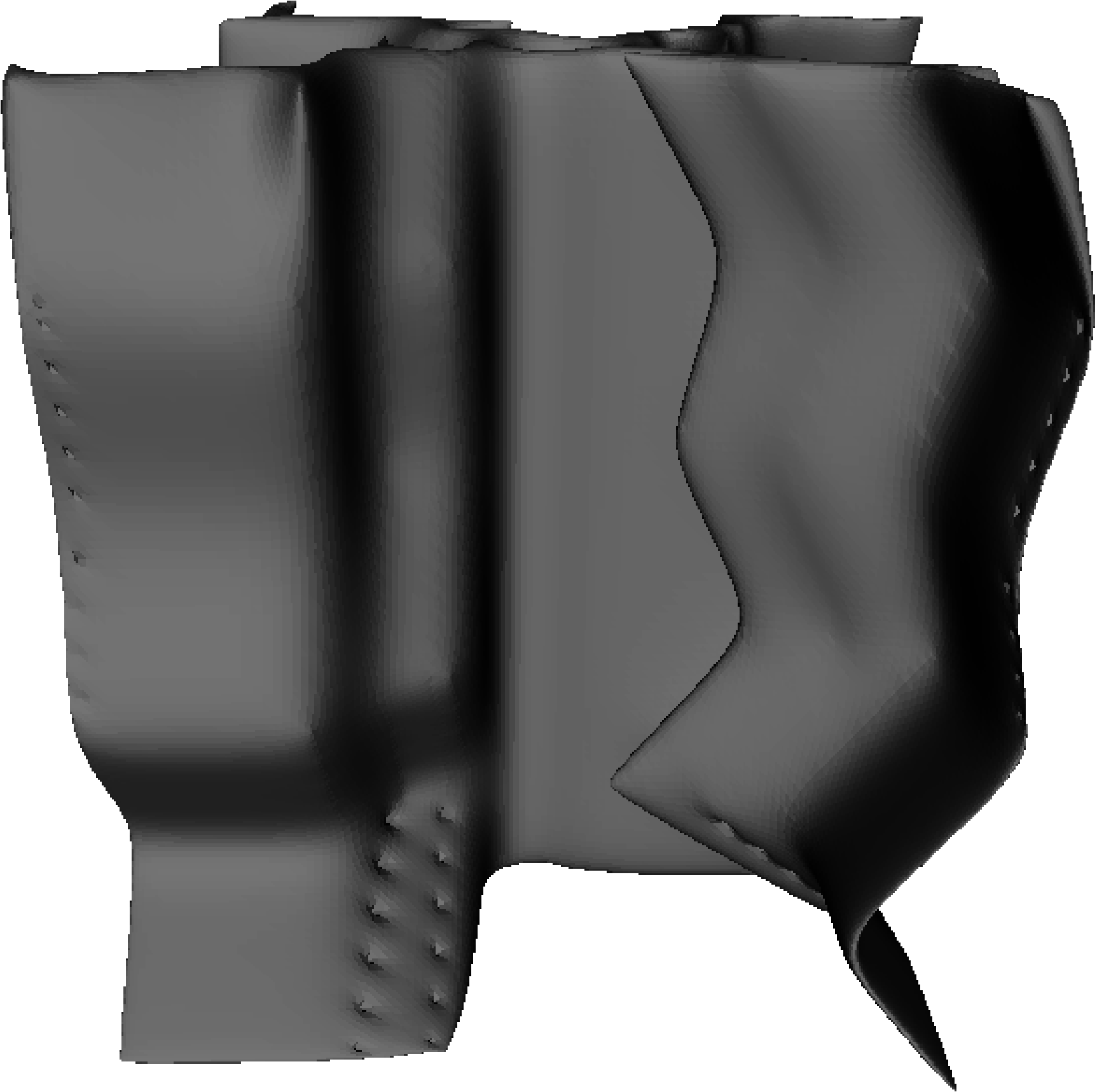}
	\caption{SCGA-ELS fittest evolved interacting VAWTs after 160 fabrications; 2692~rpm.}
	\label{fig:els-best-evolved}
\end{figure}

The average rotational speed of SCGA-ELS ($M=2264$, $SD=322$, $N=40$) is significantly greater than SCGA ($M=2112$, $SD=307$, $N=40$) using a two-tailed Mann-Whitney test, $U=469$, $p\le.0015$. Furthermore, the fittest SCGA-ELS array pairing (2692~rpm; see Figure~\ref{fig:els-best-evolved}) was greater than SCGA (2429~rpm) after 160 fabrications. However, this comparison may not be fair to the original SCGA, which fabricates a randomly selected individual as well as the best each epoch, whereas SCGA-ELS only fabricates the best. Therefore, a two-tailed Mann-Whitney test was performed for the rotational speed of SCGA-ELS over the first 20 model suggested offspring ($M=2303$, $SD=134$, $N=20$) with only the suggested best offspring from SCGA ($M=2087$, $SD=386$, $N=20$) and the result was found to be statistically significant, $U=114$, $p\le.02$. The same comparison was made for SCGA-ELS over the first 20 model suggested offspring with only the suggested best offspring from SCGA-20T ($M=1907$, $SD=443$, $N=20$) and the result was also statistically significant, $U=73.5$, $p\le.00063$. SCGA-ELS therefore seems to offer a clear benefit in these experiments.
 
\subsection{Collaboration Strategies}

Whilst it is relatively costly to fabricate a new VAWT prototype ($\sim30$~minutes), the rotational speed measurement with the laser tachometer is relatively cheap ($\le$~1~minute). Therefore, while it may not be the case for more elaborate wind tunnel testing conditions, here reevaluating some fabricated individuals with other partners may provide a relatively cheap way to increase performance. To see whether a two-partner strategy can increase the performance in designing two interacting VAWT, the CGA was rerun with the same initial population as before, however collaboration was performed with an individual that was fitness proportionately selected (i.e., roulette wheel) in addition to the usual elite member from the other species population, and the larger of the two fitness scores assigned to the individual being evaluated (CGA-2). Since initially no fitness scores are known, each individual in the first generation of a species population was evaluated in conjunction with a single randomly selected individual from the other population (that is, as before) and also with a second randomly selected individual, uniquely chosen for each evaluation, and the larger of the two shared fitness scores assigned.

So far, the CGA has assumed that there is a significant degree of asymmetry within the array solution and hence no cross-species gene sharing has been permitted. However, it is possible that there may be sufficient symmetry that can be exploited to increase performance. Therefore, the CGA was also rerun with the same initial population as before, however an offspring is also evaluated in the alternate physical position and in collaboration with the elite member from its own species population (CGA-CROSS). If the fitness is greater than the population best fitness, it is also added to that species population, replacing the worst individual. 

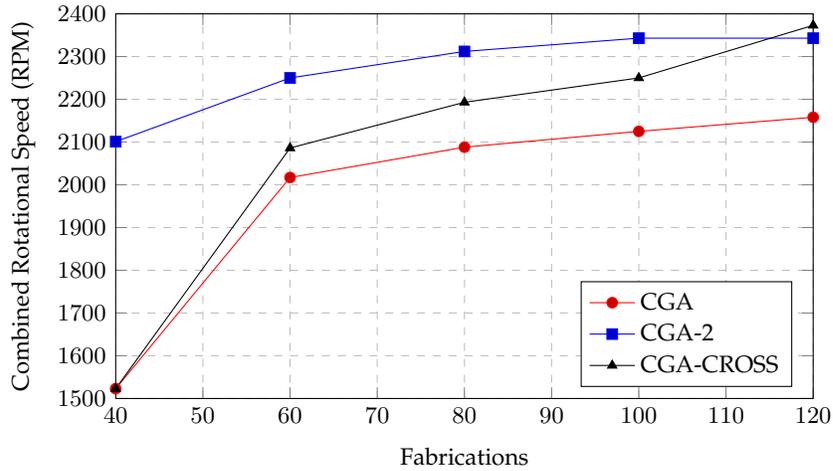
\begin{figure}[t]
	\centering 
	\begin{tikzpicture}[font=\figfontsize] 
		\begin{axis}[
				width=\graphwidth,
				height=\graphheight,
				/pgf/number format/.cd,
				1000 sep={},
				xlabel=Fabrications,
				ylabel=Combined Rotational Speed (RPM),
				grid=major,
				grid style={dashed, gray!50},
				ymin=1500,
				ymax=2400,
				ytick={1500,1600,...,2400},
				xmin=40,
				xmax=120,
				legend entries={CGA,CGA-2,CGA-CROSS},
				legend style={nodes=right},
				legend pos= south east,
				cycle list name=mark list,
			]
			\addplot+[red] table [x=a, y=b, col sep=comma] {coevo-evals.dat};
			\addplot+[blue] table [x=a, y=f, col sep=comma] {coevo-evals.dat};
			\addplot+[black] table [x=a, y=g, col sep=comma] {coevo-evals.dat};
		\end{axis}
	\end{tikzpicture}
	\caption{Array rotational speed-based evolution. Fittest array pairs. CGA (circle), CGA-2 (square), CGA-CROSS (triangle).}
	\label{fig:res-2partner}
\end{figure}

Figure~\ref{fig:res-2partner} shows the rotational speed of the fittest array pairings evolved by the original CGA, CGA-2, and CGA-CROSS over 120 fabrications. As can be seen, the fittest CGA-2 array pairing after 120 fabrications (2343~rpm) is greater than CGA (2158~rpm). Furthermore, the average combined rotational speed of the final 40 offspring formed by CGA-2 ($M=2053$, $SD=295$, $N=40$) is significantly greater than the original CGA ($M=1882$, $SD=316$, $N=40$) using a two-tailed Mann-Whitney test, $U=481$, $p\le.0022$. The extra evaluation during initialisation was used 38 out of 40 times, which provided an initial performance boost. However, during steady state evolution the extra evaluation was only used 4 out of 80 times. That is, the roulette wheel selected partner resulted in a combined rotational speed greater than when paired with the best individual only 5\% of the time. Moreover, the average difference of those fitness scores was only 12~rpm, which probably would have made little difference to the GA. Whilst not used for fitness determination during the experiment, a randomly selected partner was also evaluated with each offspring and found to have a near identical effect to the roulette wheel collaborator, being greater than the best partner 4 out of 80 times with an average difference of 46~rpm. Thus it appears that performing extra evaluations during initialisation provides a significant boost to performance over only partnering with a single random collaborator, but thereafter there is little performance benefit from performing extra evaluations.

\begin{figure}[t]
	\centering 
	\subfigure[CGA-2; 2343~rpm.] {
		\includegraphics[width=\figwidth]{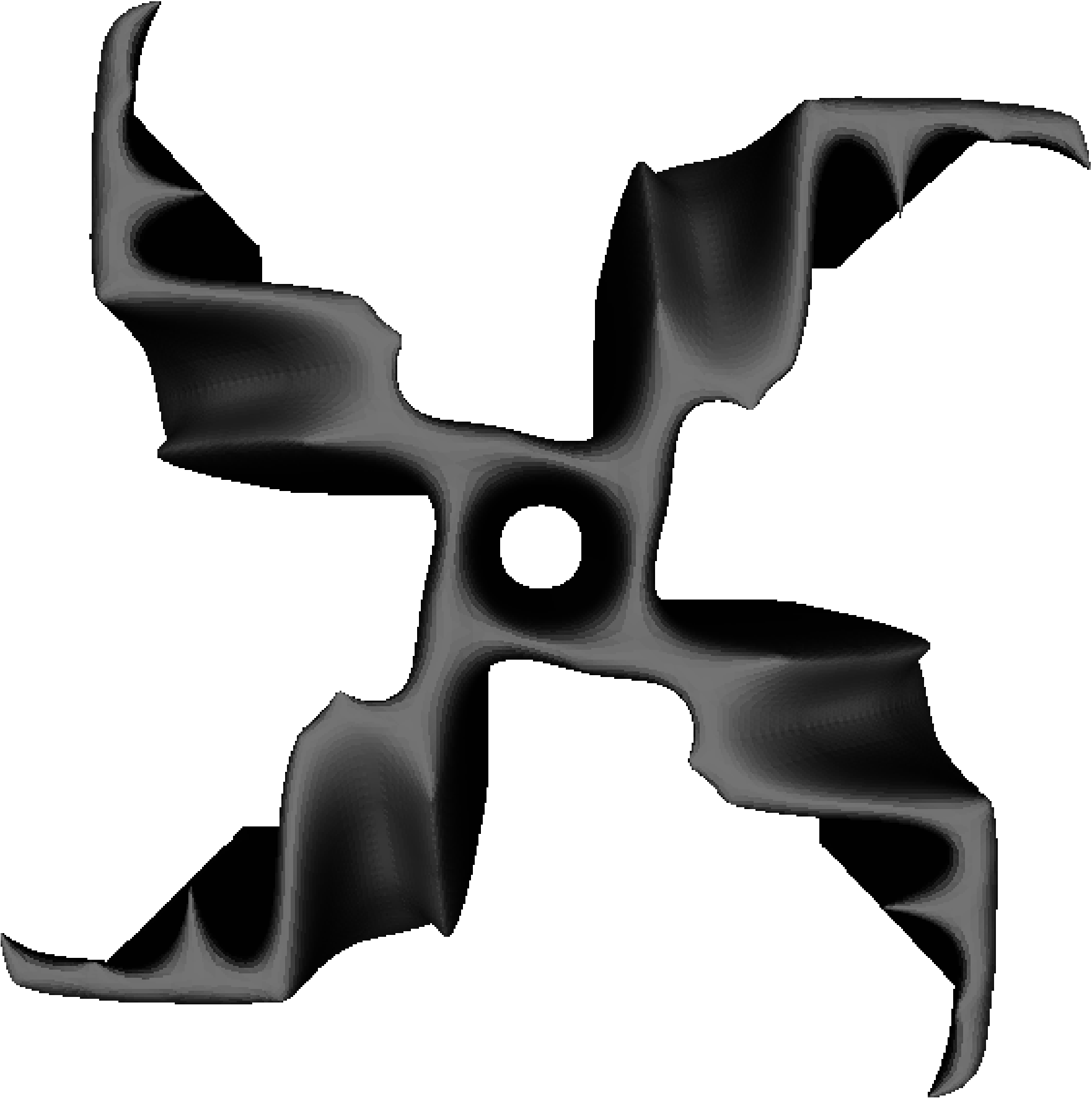} \includegraphics[width=\figwidth]{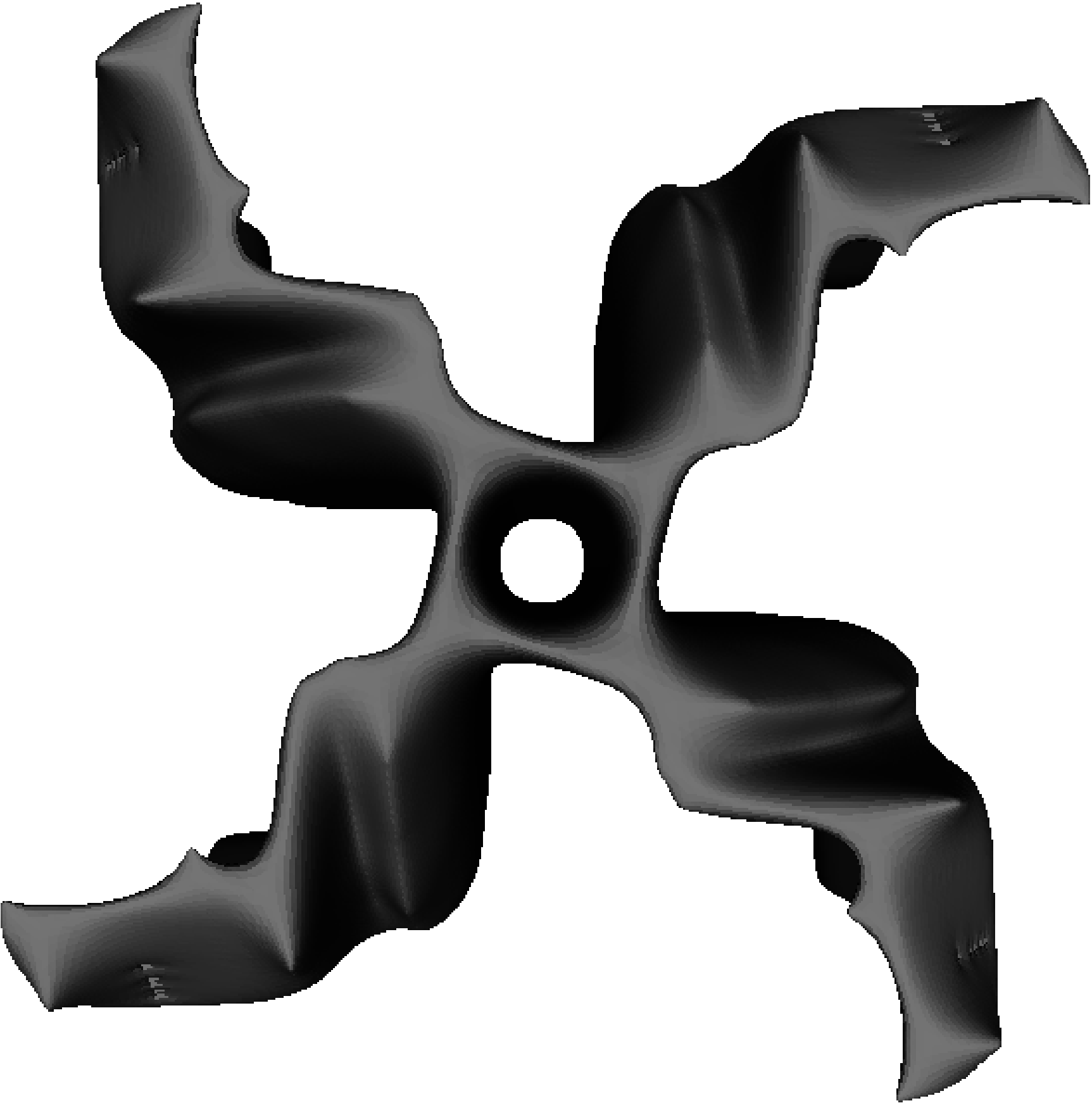} \hspace{0.3in}
		\includegraphics[width=\figwidth]{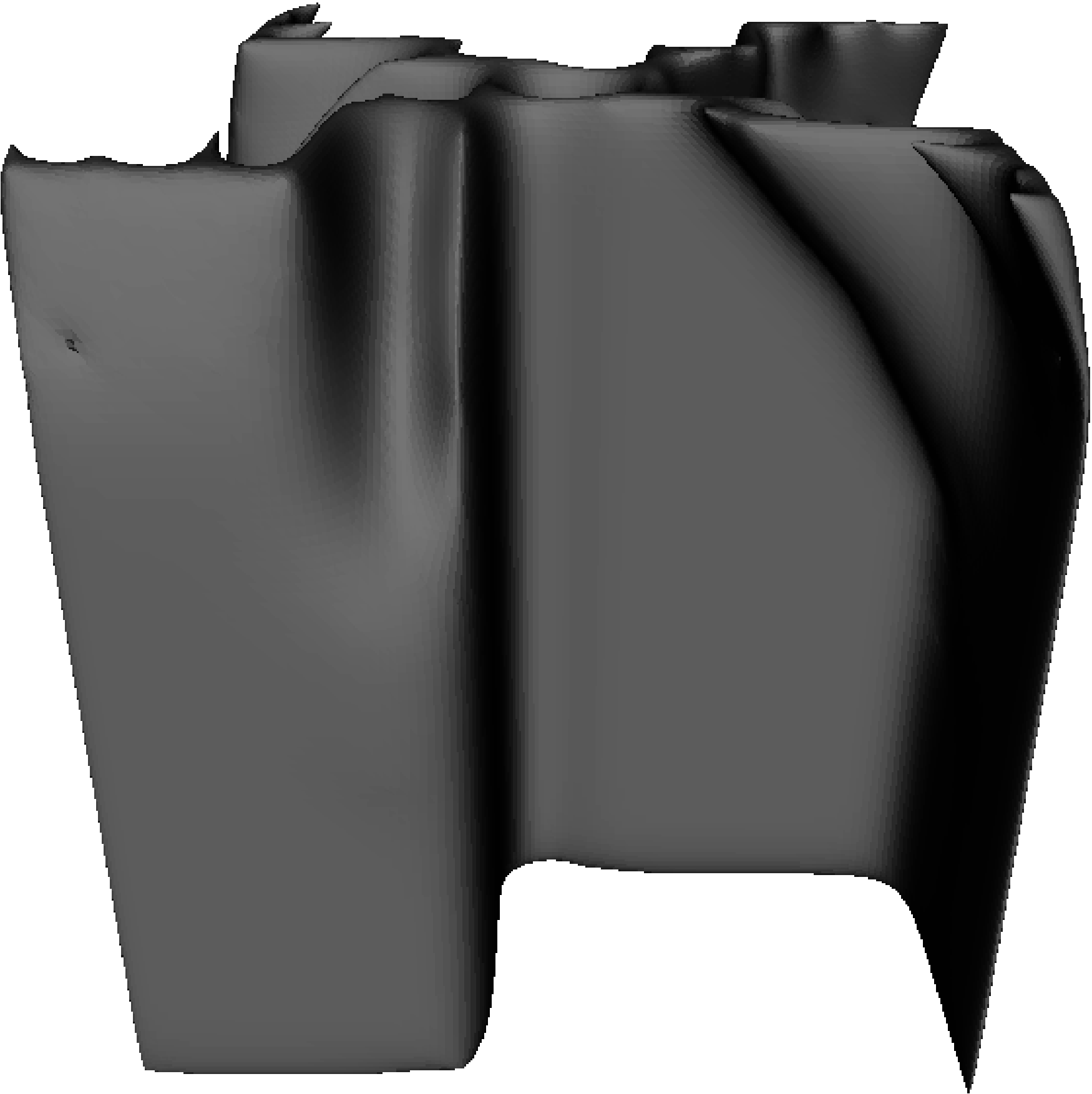} \includegraphics[width=\figwidth]{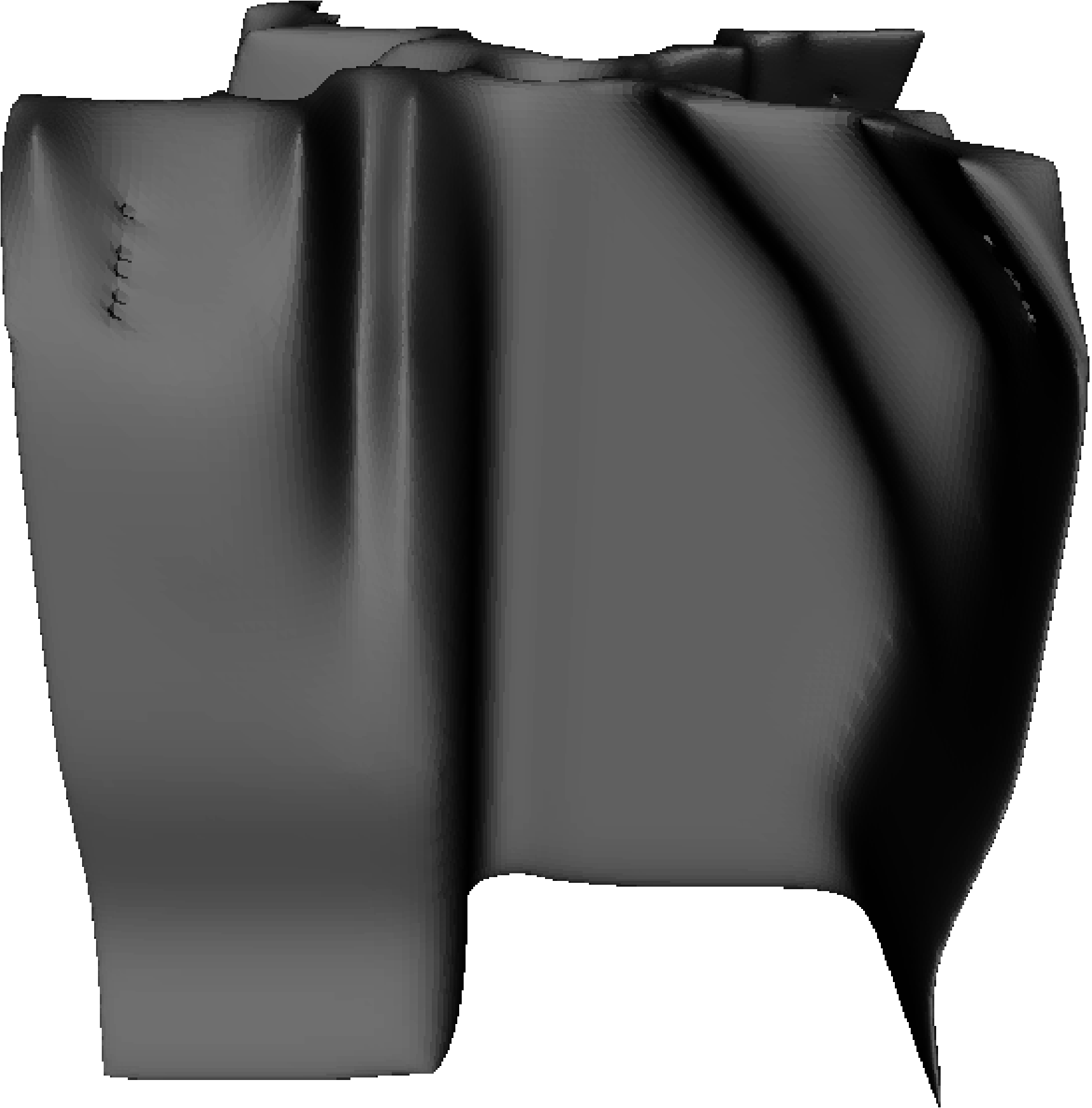} }
	\subfigure[CGA-CROSS; 2373~rpm.] {
		\includegraphics[width=\figwidth]{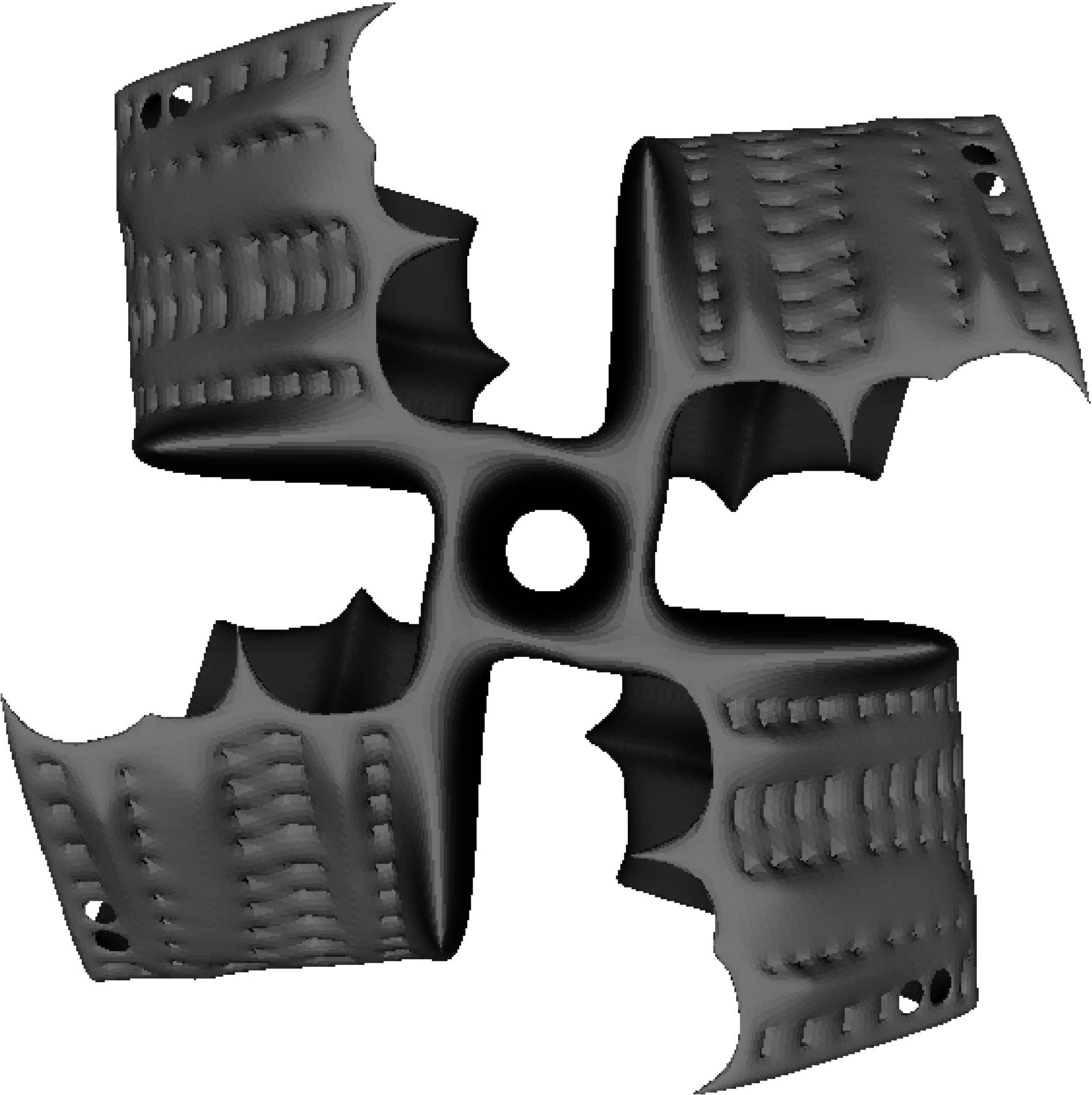} \includegraphics[width=\figwidth]{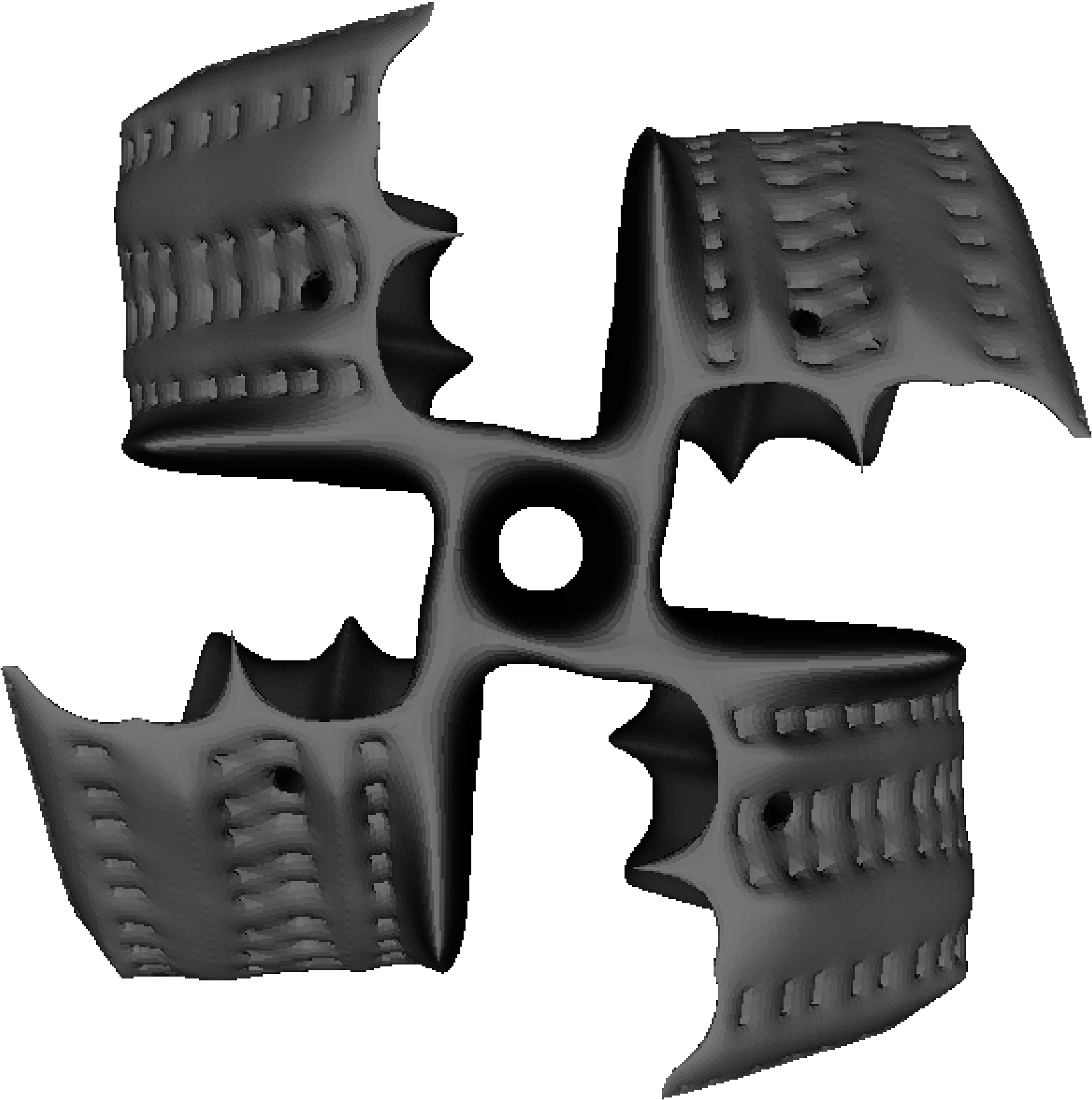} \hspace{0.3in}
		\includegraphics[width=\figwidth]{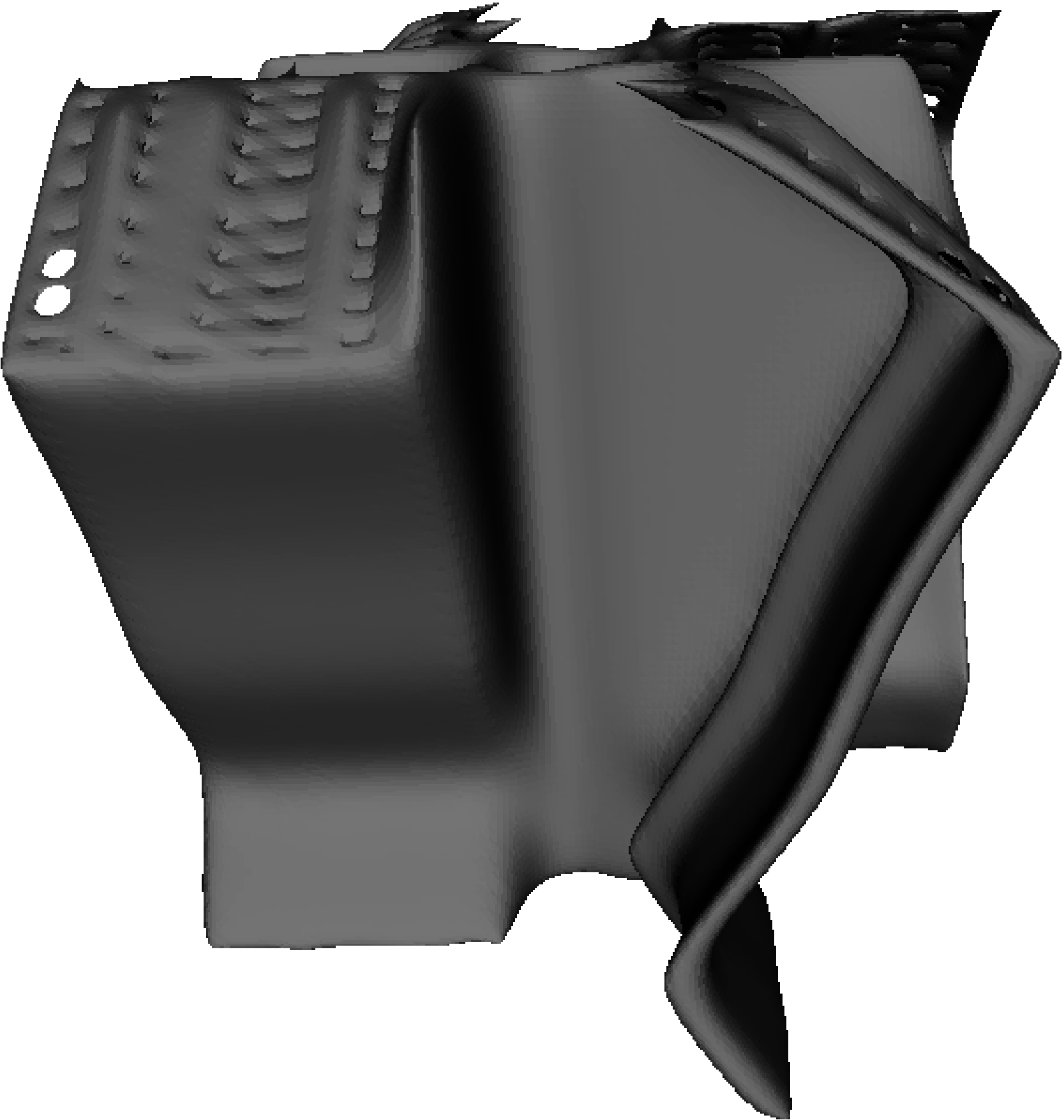} \includegraphics[width=\figwidth]{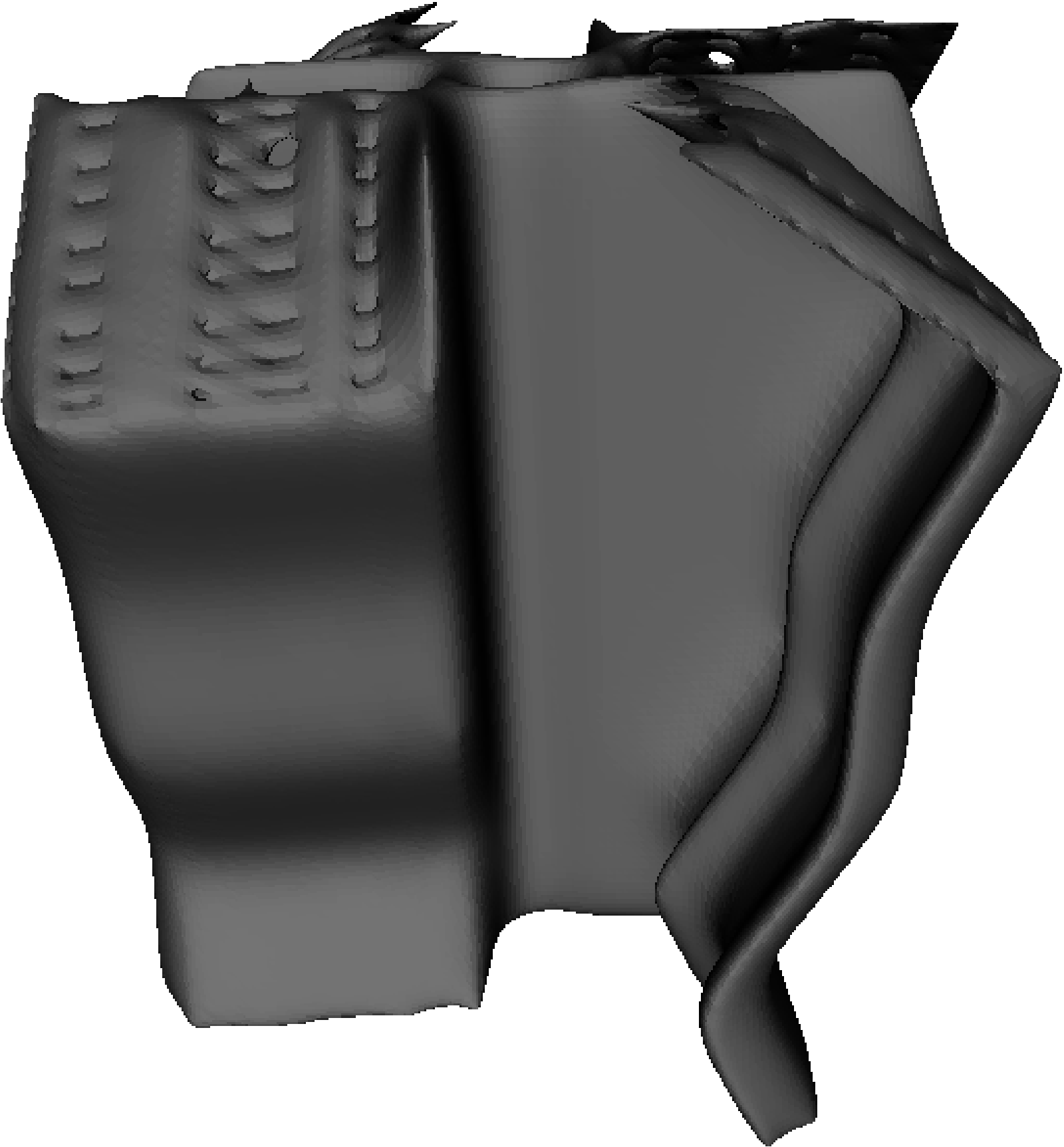} }
	\caption{Fittest evolved interacting VAWTs after 120 fabrications.}
	\label{fig:collab-best-evolved}
\end{figure}

The fittest CGA-CROSS array pairing after 120 fabrications (2373~rpm) is similar to CGA-2 (2343~rpm). In 7 out of the 80 extra evaluations performed in the CGA-CROSS experiment, evaluating the offspring in the alternate position resulted in a faster combined rotational speed than the fittest individual in that species. Furthermore, both individuals in the fittest array pairing were offspring produced in the alternate species, showing that there is a high degree of symmetry in the task. However, despite the extra evaluations having resulted in a VAWT pair with a higher rotational speed than the fittest pair from the original CGA (2158~rpm), the average combined rotational speed of the final 40 CGA-CROSS offspring evaluated in the original species position ($M=1988$, $SD=261$, $N=40$) is not statistically different to CGA using a two-tailed Mann-Whitney test, $U=655$, $p\le.163$. The fittest CGA-2 and CGA-CROSS interacting VAWTs after 120 fabrications are shown in Figure~\ref{fig:collab-best-evolved}.

\section{Conclusions}

This paper has explored a range of surrogate modelling and evolutionary techniques used to design interacting VAWTs wherein candidate prototypes are fabricated by a 3D printer and evaluated under fan generated wind conditions. The experiments here have shown that it is possible to use SCGAs to iteratively increase the performance of closely positioned Savonius turbines, exploiting the inter-turbine flow effects, which is extremely difficult to achieve under simulation. The SCGA represents a scalable approach to the design of wind turbine arrays since the number of inputs to the surrogate-models remains constant regardless of the number of turbines undergoing evolution. 

The accuracy of various modelling algorithms used to estimate the fitness of evaluated individuals from the initial experiments was compared, finding that there is little difference between the algorithms for the current task. The effect of temporally windowing the surrogate model training samples was shown to be a promising approach, however resulted in no performance benefit in practice. Subsequently, a SCGA based on an enhanced local search was introduced and found to produce more efficient designs within the same number of fabrications when compared with the original SCGA approaches. Finally, alternative coevolution collaboration schemes were examined, finding an initial performance increase resulting from more accurate fitness assignments but with no significant improvement thereafter. This highlights that while much can be learnt by developing optimisation algorithms using simulated problems and mining data, significant differences in performance are often found when applying the results to the real application. Thus, whilst general learning algorithms provide a starting point for exploration, there is no escape from further development directly in the problem application.

The use of 3D printing to physically instantiate candidate designs completely avoids the use of 3D computer simulations, with their associated processing costs and modelling assumptions. In this case, 3D CFD analysis was avoided, but the approach is equally applicable to other real-world optimisation problems, for example, those requiring computational structural dynamics or computational electromagnetics simulations. We anticipate that in the future such `design mining' approaches will yield unusual yet highly efficient designs that exploit characteristics of the environment and/or materials that are difficult to capture formally or in simulation. This has the potential to place knowledge discovery at the core of engineering design, particularly within an iterative framework such as in agile approaches.

Although the print resolution used here to build the prototypes was set at the printer default, the resolution can be adjusted to provide coarser designs at a faster rate for preliminary studies (e.g.,\ for early evolutionary candidates), or slower higher resolution prints for more subtle optimisation. Further, only PLA plastic was used here to fabricate designs, but other materials such as flexible rubbers or multi-material designs can be constructed and explored by an EA. Thus 3D printing offers a range of ways to customise the evolutionary instantiation to the design task. Multiple 3D printers can also easily be used to perform parallel fabrication to speed up the process.

Future work will include the use of the power generated as the fitness computation under various wind tunnel conditions; the coevolution of larger arrays, including the turbine positioning; the exploration of more advanced assisted learning systems to reduce the number of fabrications required; examination of the effect of seeding the population with a given design; investigation of alternative representations that provide more flexible designs including variable number of blades, for example, supershapes~\citep{PreenBull:2014b}; and the production of 1:1 scale designs.

The issue of scalability remains an important future area of research. When increasing the scale of designs it is widely known that the changes in dimensionality will greatly affect performance, however it remains to be seen how performance will change in the presence of other significant factors such as turbine wake interactions in the case of arrays. One potential solution is to simply use larger 3D printing and wind-tunnel capabilities whereby larger designs could be produced by the same method. On the opposite end of the spectrum, micro-wind turbines that are 2~mm in diameter or smaller can be used to generate power, such as for wireless sensors~\citep{Howey:2011}, and in this case more precise 3D printers would be required. Moreover, wind turbines can find useful applications on any scale, e.g.,\ a recent feasibility study~\citep{Park:2012} for powering wireless sensors on cable-stayed bridges examined turbines with a rotor diameter of 138~mm in wind conditions with an average of 4.4~m/s (similar to the artificial wind conditions used in this paper).

If the recent speed and material advances in rapid-prototyping continues, along with the current advancement of evolutionary design, it will soon be feasible to perform a wide-array of automated complex engineering optimisation {\it in situ}, whether on the micro-scale (e.g.,\ drug design), or the macro-scale (e.g.,\ wind turbine design). That is, instead of using mass manufactured designs, EAs will be used to identify bespoke solutions that are manufactured to compensate and exploit the specific characteristics of the environment in which they are deployed, e.g.,\ local wind conditions, nearby obstacles, and local acoustic and visual requirements for wind turbines. 
 
\acknowledgments
\\This work was supported by the UK Leverhulme Trust under Grant RPG-2013-344.

\small
%

\end{document}